\RequirePackage[l2tabu, orthodox]{nag}

\documentclass[12pt,mphil,a4paper,oneside]{ucl_thesis}
\usepackage{amsmath}
\usepackage{graphicx} 
\usepackage{algorithm}
\usepackage{algpseudocode}
\usepackage{amssymb}
\usepackage{amsmath, amssymb, amsthm}
\usepackage{xcolor}
\usepackage{enumitem}

\usepackage{blindtext}

\usepackage{emptypage}

\usepackage{graphicx}

\usepackage{float}

\usepackage{amsmath}

\usepackage{gensymb}
\usepackage{textcomp}

\usepackage{setspace}
\usepackage{multirow}

\usepackage{bibentry} 

\usepackage[format=hang,font=small,labelfont=bf]{caption}

\usepackage{etoolbox}

\usepackage{microtype}

\usepackage{bibentry}
\makeatletter\let\saved@bibitem\@bibitem\makeatother
\usepackage[pdftex,hidelinks]{hyperref}
\makeatletter\let\@bibitem\saved@bibitem\makeatother
\makeatletter
\AtBeginDocument{
    \hypersetup{
        pdfsubject={Thesis Subject},
        pdfkeywords={Thesis Keywords},
        pdfauthor={Author},
        pdftitle={Title},
    }
}
\makeatother

\begin{document}

\nobibliography*



\title{Optimizing Medication Decisions for Patients with Atrial Fibrillation through Path Development Network}
\author{Tian Xie 22196037}
\department{Department of Mathematics}

\maketitle

\begin{abstract} 
Atrial fibrillation (AF) is a common cardiac arrhythmia characterized by rapid and irregular contractions of the atria. It significantly elevates the risk of strokes due to slowed blood flow in the atria, especially in the left atrial appendage, which is prone to blood clot formation. Such clots can migrate into cerebral arteries, leading to ischemic stroke. To assess whether AF patients should be prescribed anticoagulants, doctors often use the CHA2DS2-VASc scoring system. However, anticoagulant use must be approached with caution as it can impact clotting functions. This study introduces a machine learning algorithm that predicts whether patients with AF should be recommended anticoagulant therapy using 12-lead ECG data. In this model, we use STOME to enhance time-series data and then process it through a Convolutional Neural Network (CNN). By incorporating a path development layer, the model achieves a specificity of 30.6\% under the condition of an NPV of 1. In contrast, LSTM algorithms without path development yield a specificity of only 2.7\% under the same NPV condition.

\end{abstract}

\begin{impactstatement}

\subsection*{Impact Within Academia}

\noindent \textbf{Academic Research and Future Scholarship}

Utilizing the path development layer in machine learning models not only carves a new path of research in the arena of arrhythmia and stroke risk prediction but also promotes interdisciplinary studies including the intersection of machine learning and medical science.

\noindent \textbf{Research Methods or Methodology}

The study not only validates the superiority of the path development layer in feature extraction but also showcases its potential to work synergistically with Long Short-Term Memory (LSTM) networks, offering new avenues and opportunities for future research.

\subsection*{Impact Outside Academia}

\noindent \textbf{Public Health and Policy}

The research presents a solution with the potential to significantly enhance the treatment methodology for patients with atrial fibrillation, reducing reliance on anticoagulants, thereby decreasing the associated health risks and medical costs.

\noindent \textbf{Professional Practice}

Medical professionals can leverage this new technology for a better assessment of stroke risk in patients, moving beyond the current methodologies, potentially leading to more personalized treatment plans.

\noindent \textbf{Quality of Life}

By diminishing the utilization of anticoagulants, we can aid patients in avoiding the side effects associated with these medications, thereby improving their quality of life.

\end{impactstatement}

\setcounter{tocdepth}{2} 

\tableofcontents
\listoffigures
\listoftables

\chapter{Introduction}
\label{chapterlabel1}

Atrial fibrillation (AF) is a common cardiac arrhythmia characterized by rapid and irregular contractions of the atria\cite{silverman1994rebellious}. It is closely related to stroke, also known as cerebral infarction. Under atrial fibrillation, the atria no longer contract effectively, leading to a slowdown in blood flow within the atria, which increases the risk of clot formation\cite{ezekowitz1999preventing}. Blood clots are especially prone to form in the left atrial appendage, a small, pouch-like structure within the atrium. Once clots form in the atria, they may be carried into the ventricles and then into the systemic circulation\cite{silverman1994rebellious}. If a clot enters an artery supplying the brain, it could block that artery, leading to an ischemic stroke\cite{wolf1991atrial}. To assess and reduce the stroke risk in AF patients, doctors often use tools like the CHA2DS2-VASc scoring system\cite{gazova2019predictive}. This system evaluates the patient's stroke risk based on several factors, such as age, gender, presence of hypertension, or diabetes\cite{gazova2019predictive}. For AF patients at higher risk, doctors typically prescribe anticoagulants, like warfarin or newer oral anticoagulants (NOACs), such as dabigatran or rivaroxaban, to reduce the risk of stroke\cite{gage2000adverse}. However, these drugs also reduce the patient's blood clotting ability, so they should not be used in patients with extremely low stroke probabilities\cite{gage2000adverse}.

Electrocardiography stands as an integral pillar in the realm of cardiovascular diagnostics, particularly for conditions like atrial fibrillation (AF)\cite{sodmann2018convolutional}. The Electrocardiogram (ECG), a quintessential embodiment of this method, captures digital signals emanating from the heart's electric flux, achieved by placing electrodes strategically over the human physique \cite{hao2020multibranch}. This allows for the discernment of the heart's voltage fluctuations, offering a direct window into its rhythmic behavior.

The conventional modus operandi has cardiologists painstakingly scrutinizing waveforms on 12-lead ECGs, presented as digital imagery. This task grows exponentially laborious when faced with ECG traces that span extensive durations, sometimes stretching over multiple days. Such a meticulous, protracted process not only drains the expert's time and energy but also introduces the potential for subjective bias, detracting from the diagnosis's objectivity.
With the advent of advanced computational techniques, there has been a surge in the development of machine learning-based algorithms tailored for automated electrocardiogram (ECG) waveform analysis. Nonetheless, it is imperative to note that the efficacy of these algorithms is profoundly influenced by factors such as the sampling rate and electrode placement points, potentially affecting the accuracy and reliability of the derived results\cite{sodmann2018convolutional}\cite{hao2020multibranch}\cite{tadesse2019cardiovascular}.
In contemporary academic discourse, there has been a marked surge in the development and application of machine learning algorithms tailored for the analysis of electrocardiograms (ECGs). A plethora of methodologies encompassing Support Vector Machines (SVM) and Genetic Algorithm (GA)\cite{khazaee2013heart}, Deep Learning architectures\cite{maglaveras1998adaptive}, Transfer Learning techniques\cite{tadesse2019cardiovascular}, Convolutional Neural Networks (CNN)\cite{sodmann2018convolutional}, Principal Component Analysis (PCA)\cite{maglaveras1998ecg}, Long Short-Term Memory networks (LSTM)\cite{gao2019effective}, as well as hybrid configurations of CNN and LSTM have been brought to the forefront\cite{petmezas2021automated}. Despite their promising capabilities, these computational paradigms confront inherent limitations. One of the salient challenges pertains to their substantive demand for data and computational memory resources. This requirement becomes particularly exigent considering the constraints associated with the accessibility and volume of medical datasets, a concern that is amplified within the confines of the healthcare sector.

In the field of machine learning, classification tasks involve assigning an input instance from an input space \( \mathcal{X} \) to one of the predefined categories in an output space \( \mathcal{Y} \). In the context of image classification, \( \mathcal{X} \) encompasses all conceivable images, typically represented as high-dimensional vectors, where each component indicates the brightness or RGB value of a pixel. For classification, the output space \( \mathcal{Y} \) is a finite and discrete set of labels. Taking a canine-feline dichotomy as an example, \( \mathcal{Y} \) would be \{\text{"cat"}, \text{"dog"}\}. There's also a hypothesis space \( \mathcal{H} \), which constitutes all possible mappings (functions) from \( \mathcal{X} \) to \( \mathcal{Y} \). Model training can be perceived as a search within \( \mathcal{H} \) to ascertain the optimal function that predicts the accurate output for a given input, based on the available training dataset. A loss function is a scalar function, denoted as \( L: \mathcal{Y} \times \mathcal{Y} \rightarrow \mathbb{R} \), used to quantify the disparity between the predicted label and the true label. In the context of classification, the Cross-Entropy Loss is ubiquitously employed\cite{krizhevsky2012imagenet}. Given a training dataset \( \mathcal{D} = \{(x_1, y_1), (x_2, y_2), \ldots, (x_n, y_n)\} \) where \( x_i \in \mathcal{X} \) and \( y_i \in \mathcal{Y} \), the goal is to deduce a function \( f \in \mathcal{H} \) such that the prediction \( \hat{y}_i = f(x_i) \) aligns with the true label \( y_i \). The empirical risk, based on the loss function, can be formulated as \[ R(f) = \frac{1}{n} \sum_{i=1}^{n} L(y_i, f(x_i)) \]The essence of the training phase is to identify a hypothesis \( f^* \in \mathcal{H} \) that minimizes this empirical risk over \( \mathcal{D} \).

The evolution of neural network algorithms spans several decades, encompassing numerous theoretical and technological breakthroughs. Frank Rosenblatt first introduced the Perceptron model in 1957, laying the groundwork for neural network research\cite{rosenblatt1957perceptron}. However, in 1969, Minsky and Papert's work "Perceptrons" pointed out the limitations of single-layer perceptrons in handling nonlinear problems\cite{minsky1969perceptrons}. This limitation inspired research on multi-layer neural networks. By the mid-1980s, the integration of multi-layer feedforward networks with the backpropagation algorithm became mainstream\cite{rumelhart1986learning}. Rumelhart, Hinton, and Williams introduced the backpropagation algorithm in 1986, making the training of multi-layer neural networks feasible.

Subsequently, to deal with data of a sequential nature, Recurrent Neural Networks (RNNs) were proposed\cite{hochreiter1997long}. This network structure can remember previous inputs, making it highly effective for temporal data like time series and natural language. However, traditional RNNs suffer from the vanishing and exploding gradient problems, limiting their ability to capture long-term dependencies. To address these issues, Hochreiter and Schmidhuber proposed the Long Short-Term Memory (LSTM) network in 1997, which effectively captures long-term sequence dependencies through gating mechanisms\cite{hochreiter1997long}.

Further advancements include Yann LeCun's introduction of LeNet-5 in 1998, an early convolutional neural network for handwritten digit recognition\cite{lecun1998gradient}. In 2012, AlexNet, developed by Krizhevsky, Sutskever, and Hinton, achieved significant success in the ImageNet challenge, marking the advent of the deep learning era\cite{krizhevsky2012imagenet}. Recent research also explores new technologies such as residual networks\cite{he2016deep}, generative adversarial networks\cite{goodfellow2014generative}, and the interpretability and robustness of neural networks\cite{mangal2019robustness}.

Path development is a newly proposed machine learning algorithm for time series processing that can extract features using Lie algebra\cite{lou2022path}. Using this strategy, one can map paths to Lie group sequences, accomplishing feature extraction for long time series.


\chapter{Prerequisites on Machine Learning}
\label{chapterlabel2}

The main focus here is on neural networks, while general machine learning content is in the appendix.
\section{Neural networks}

Neural networks are computational models that simulate the way the human brain works, used for pattern recognition and classification\cite{rumelhart1986learning}. They consist of multiple nodes (or "neurons") that are interconnected across layers. Each connection has a weight, representing the importance of that connection.A typical neural network consists of three types of layers: input layer, hidden layers, and output layer. Data starts at the input layer, passes through the hidden layers, and finally reaches the output layer. In a multi-layer perceptron (MLP), each neuron of one layer is fully connected to every neuron of the next layer.
In a neural network, the transformation from the hidden layers to the output can be described mathematically. 

In a neural network, the transformation from the hidden layers to the output can be described mathematically. 

For the hidden layers, the output vector \(\mathbf{y_h}\) can be derived using:

\begin{equation}
\mathbf{y_h} = \sigma(\mathbf{W_h} \cdot \mathbf{x} + \mathbf{b_h})
\end{equation}

Where:

\begin{itemize}
    \item \( \mathbf{x} \) represents the output vector from the previous layer or the input layer.
    \item \( \mathbf{W_h} \) is the weight matrix for that hidden layer.
    \item \( \mathbf{b_h} \) denotes the bias vector for the hidden layer.
    \item \( \sigma \) is a non-linear activation function, with common choices being the Rectified Linear Unit (ReLU) or the sigmoid function.
    \item \( \mathbf{y_h} \) is the output vector of this hidden layer, which will also serve as the input for the subsequent layer or the output layer.
\end{itemize}

For the output layer, the output vector \(\mathbf{y_o}\) is given by:

\begin{equation}
\mathbf{y_o} = \sigma_o(\mathbf{W_o} \cdot \mathbf{y_h} + \mathbf{b_o})
\end{equation}

Where:

\begin{itemize}
    \item \( \mathbf{y_h} \) is the output vector from the last hidden layer.
    \item \( \mathbf{W_o} \) represents the weight matrix for the output layer.
    \item \( \mathbf{b_o} \) denotes the bias vector for the output layer.
    \item \( \sigma_o \) is typically a softmax function for classification tasks or another suitable activation function depending on the task.
    \item \( \mathbf{y_o} \) is the final output of the network
    \begin{equation}
    \mathbf{y_o} = \sigma_o(\mathbf{W_o} \cdot  \sigma(\mathbf{W_h} \cdot \mathbf{x} + \mathbf{b_h}) + \mathbf{b_o})
    \end{equation}
\end{itemize}

To train a neural network, we use an algorithm called backpropagation, combined with gradient descent, to optimize the weights and minimize the discrepancy between predicted and actual values\cite{rumelhart1986learning}. The loss function \( L \) quantifies this discrepancy, for instance, mean squared error or cross-entropy loss.

For a given training dataset \( \mathcal{D} = \{(x_1, y_1), (x_2, y_2), \ldots, (x_n, y_n)\} \), the goal is to find a vector \( \mathbf{W_h^*} \),\( \mathbf{b_h^*} \),\( \mathbf{W_o^*} \),\( \mathbf{b_o^*} \)  that minimizes the loss function \( L \).

The process of training a neural network is iterative, with each iteration updating the weights to progressively improve the network's performance\cite{rumelhart1986learning}. Once trained, the network can be used to predict new, unseen data.

A salient inquiry pertains to the capability of such neural architectures to serve as universal function approximators. The universal approximation theorem posits that, given certain prerequisites, neural networks possess the potential to approximate any continuous function to an arbitrary degree of accuracy.

\newtheorem{theorem}{Theorem}

\begin{theorem}[Cybenko, 1989]\cite{cybenko1989approximation}
For a standard sigmoid activation function \( \sigma \) (or other squashing functions), for any continuous function \( f: [0, 1]^m \to \mathbb{R} \) and any \( \epsilon > 0 \), there exists a neural network \( N_w \) such that:
\[
\max_{x \in [0, 1]^m} |f(x) - N_w(x)| < \epsilon
\]
where
\[
N_w(x) = \sum_{j=1}^N \alpha_j \sigma(\mathbf{w}_j \cdot \mathbf{x} + b_j).
\]
\end{theorem}

\begin{theorem}[Hornik, 1991]\cite{hornik1989multilayer}
For any non-constant, bounded, monotonically-increasing activation function \( \sigma \), and for any continuous function \( f: [0, 1]^m \to \mathbb{R} \), for any \( \epsilon > 0 \), there exists a neural network \( N_w \) such that:
\[
\max_{x \in [0, 1]^m} |f(x) - N_w(x)| < \epsilon
\]
with the same form of \( N_w(x) \) as in the Cybenko theorem.
\end{theorem}

\begin{theorem}[Pinkus, 1999]\cite{pinkus1999approximation}
For any non-algebraic polynomial activation function \( \sigma \),Let \( K\subset \mathbb{R}^m\) be compact, and for any continuous function \( f: K \to \mathbb{R} \), for any \( \epsilon > 0 \), there exists a neural network \( N_w \) such that:
\[
\max_{x \in K} |f(x) - N_w(x)| < \epsilon
\]
with the same form of \( N_w(x) \) as in the previous theorems.
\end{theorem}

Upon a modest generalization, one can derive the approximation theorem pertinent to neural networks with a singular hidden layer.

\begin{theorem}
For any non-polynomial activation function \( \sigma \),Let \( K\subset \mathbb{R}^m\) be compact, and for any continuous function \( f: K \to \mathbb{R}^n \), for any \( \epsilon > 0 \), there exists a neural network \( \mathbf{N}_w: \mathbb{R}^m \to \mathbb{R}^n \) such that:
\[
\max_{x \in K} \| f(x) - \mathbf{N}_w(x) \|_\infty < \epsilon
\]
where
\[
\mathbf{N_w}(x)_i = \sum_{j=1}^N \alpha_{ij} \sigma(\mathbf{w}_{ij} \cdot \mathbf{x} + b_{ij}),i\in\{1, 2, 3, \ldots, n\}.
\]
\end{theorem}
\(\forall k \in \mathbb{N}\), if the activation function is a polynomial of degree \( k \), then the neural network is also a polynomial of at most degree \( k \), if the set \( K \) has a non-empty interior. Polynomials of degree \( k \) are not dense in \( C(K) \) space, so the approximation theorem does not hold. The proof for this is in the appendix.


\section{Deep Neural Network, Convolutional Neural Networks}

Deep Neural Networks (DNNs) are essentially an extension of single-layer neural networks, incorporating multiple layers to extract more complex features from input data. This network does not contain just a single hidden layer but several, forming a "deep" network. By adding more hidden layers, it can learn more complex and abstract features, which is vital for tackling a range of advanced tasks and dealing with intricate datasets\cite{abiodun2018state}.

The mathematical formulation of deep neural networks can be understood with this.

\begin{itemize}
    \item \textbf{Input data}: \(\mathbf{X} = [x_1, x_2, \ldots, x_n]\)
    \item \textbf{Weights}: The weight matrix for the \(l\)-th layer is denoted as \(\mathbf{W}^{[l]}\)
    \item \textbf{Biases}: The bias vector for the \(l\)-th layer is denoted as \(\mathbf{b}^{[l]}\)
    \item \textbf{Activation functions}: Generally represented as \(\sigma^{[l]}\), it can be sigmoid, tanh, ReLU, etc.
    \item \textbf{Number of layers}: The DNN contains \(L\) layers, each harboring a certain number of neurons
\end{itemize}

In forward propagation, both the linear and activation outputs for each layer are computed. For the \(l\)-th layer, the computations are as follows:

\[
\mathbf{Z}^{[l]} = \mathbf{W}^{[l]}\mathbf{A}^{[l-1]} + \mathbf{b}^{[l]}
\]
\[
\mathbf{A}^{[l]} = \sigma^{[l]}(\mathbf{Z}^{[l]})
\]

Where:
\begin{itemize}
    \item \(\mathbf{A}^{[l]}\): The activation output of the \(l\)-th layer
    \item \(\mathbf{Z}^{[l]}\): The linear output of the \(l\)-th layer
    \item \(\mathbf{A}^{[0]}\): The input data, denoted as \(\mathbf{X}\)
\end{itemize}

\[
\hat{y} = \sigma^{[L]} \left( \mathbf{W}^{[L]} \cdot \sigma^{[L-1]}(\cdots \sigma^{[2]}(\mathbf{W}^{[2]} \cdot \sigma^{[1]}(\mathbf{W}^{[1]} \cdot \mathbf{x} + \mathbf{b}^{[1]}) + \mathbf{b}^{[2]}) \cdots ) + \mathbf{b}^{[L]} \right)
\]

Where:
\begin{itemize}
    \item \( \hat{\mathbf{y}} \) is the output vector of the network.
\end{itemize}

In this formulation, each layer applies a linear transformation using its respective weight matrix and bias vector, followed by a non-linear activation function. This series of transformations allows the DNN to learn complex hierarchical features from the input data. Nevertheless, training issues have remained a persistent concern\cite{Glorot2010}.

In the field of neural network architectures, the critical role of depth in facilitating a more parsimonious parameterization while approximating continuous functions, all the while maintaining a potent capability for feature extraction, is undeniable. Through meticulous mathematical constructions, the research illuminates the existence of specific functions that can be adeptly represented by a neural network possessing a depth of \(O(k^3)\), thereby establishing a substantive demarcation in the approximation capabilities when juxtaposed with shallower networks. Notably, to approximate these functions with a network characterized by a diminished depth, an exponentially burgeoning number of nodes are necessitated, to be precise, a lower bound delineated by \(\Omega(2^k)\)\cite{pmlr-v49-telgarsky16}. This significant discovery further reinforces the expressive power of deep neural networks in managing complex functions and patterns, especially with a notable increase in network depth\cite{raghu2017on}. It not only offers fresh insights into the design and training of neural networks, particularly in deciding the depth and breadth of the network. On the other hand, other studies highlight that width also plays a crucial role in the network's expressive power. While depth can enhance the network's expressive power, width cannot be overlooked, as it provides a potent way to increase the network's complexity and capabilities. A deep and detailed research and discussion are required in real-world applications to find the optimal balance between depth and width\cite{Lu2017TheEP}.

In deep learning tasks, neural networks learn abstract representations from input data through their multilayer structure\cite{LeCun2015}. The initial layers mainly focus on capturing simple and basic features, such as edges and basic textures in image recognition, or words and phrases in text processing. Then, at the intermediate layers, the network starts to build more complex representations by utilizing the basic features recognized in the initial layers, identifying more advanced patterns or structures through the combination of earlier recognized simple features. As we move to the subsequent layers deeper into the network, it continues this process, carrying out further feature combination and abstraction to identify even more advanced and abstract features, like entire objects or significant parts of objects\cite{10.1007/978-3-319-10590-1_53}\cite{erhan2009visualizing}. Finally, at the output layer, the network integrates all the features learned in the preceding layers to accomplish specific tasks, such as classification or regression. This hierarchical learning approach allows deep neural networks to construct a feature hierarchy from simple to complex, capturing rich and intricate patterns in the input data, which is a key factor in their success in complex tasks such as image recognition and natural language understanding\cite{jarrett2009best}.

In the field of deep learning, Convolutional Neural Networks (CNNs) stand as a distinctive category of neural networks optimized for processing grid-topology data, such as images and videos\cite{lecun1998gradient}. CNNs surpass traditional fully connected neural networks in efficiency by utilizing convolutional and pooling layers specially designed to handle such data structures.

In computer vision, CNNs empower image classification tasks, distinguishing between various objects or entities within an image, setting a benchmark in differentiating cats from dogs, and even identifying and outlining specific objects in images through object detection and segmentation techniques\cite{krizhevsky2012imagenet}\cite{simonyan2015deep}. The realm of autonomous vehicles leverages CNNs for real-time object detection and tracking to identify and monitor other vehicles, pedestrians, and obstacles on the road.

Furthermore, they play a pivotal role in the medical field, aiding in the diagnosis of diseases by scrutinizing anomalies in MRI, CT, or X-ray images, and executing image registration tasks that align medical images from different perspectives or time frames\cite{10.1007/978-3-319-24574-4_28}. In video surveillance, they assist in recognizing abnormal behaviors or particular activities within surveillance footage\cite{ding2014violence}. They are also central to facial recognition technologies utilized for identity verification, such as the facial unlock feature in smartphones.

The convolutional layer is the cornerstone of CNNs, instrumental in extracting local features from input data. This extraction is achieved by sliding a convolutional kernel or filter over the input data. A typical convolution operation is mathematically represented as:

\[
Y(x, y) = \sum_{m=-a}^{a} \sum_{n=-b}^{b} I(x-m, y-n) K(m, n)
\]

Where:
\begin{itemize}
    \item \(Y\) is the output feature map, is the input for the next layer.
    \item \(I\) represents the input image or feature map
    \item \(K\) stands for the convolutional kernel
    \item \(a, b\) are half the dimensions of the kernel
    \item \((x, y)\) denotes a point on the output feature map corresponding to a local region in the input image
\end{itemize}

Pooling layers are utilized to reduce the dimensionality of the feature maps, lowering computational complexity while retaining pivotal information. Common pooling techniques involve max pooling and average pooling:

\textbf{Max Pooling}

\[
Y(i,j) = \max_{m=i}^{i+k-1} \max_{n=j}^{j+k-1} X(m,n)
\]

\textbf{Average Pooling}

\[
Y(i,j) = \frac{1}{k^2}\sum_{m=i}^{i+k-1} \sum_{n=j}^{j+k-1} X(m,n)
\]

Here, \(k\) is the size of the pooling window.

Convolutional layers play a pivotal role in reducing the number of parameters and mitigating the risk of overfitting in Convolutional Neural Networks. This is achieved through the following means:

\begin{enumerate}
    \item \textbf{Local Connectivity}: Each neuron is connected only to a small region of the input data, significantly reducing the number of parameters.
    
    \item \textbf{Parameter Sharing}: The parameters of the convolutional kernel are shared across the entire field of view, not only reducing the number of parameters but also allowing the network to recognize the same features in different locations.
    
    \item \textbf{Sparse Interactions}: By limiting the connectivity range of neurons, sparse interactions are achieved, further reducing the number of parameters and computational complexity\cite{li2017not}.
\end{enumerate}

These factors collectively render the convolutional layer an efficient and powerful tool, particularly in handling high-dimensional input data, as it substantially reduces model complexity while maintaining performance.

In the discipline of deep learning, particularly in the context of CNNs, the concept of the receptive field is pivotal. The receptive field delineates the specific region within the input data that can be perceived or processed by a neuron, essentially defining the extent to which variations in the input data can affect the output of a given neuron\cite{lin2017feature}.

At the inception stages of a neural network, a neuron's receptive field is generally confined, directly correlating with the dimensions of the convolutional kernel employed. As we progress to the more profound layers of the network, the receptive field expands incrementally due to the cumulative effect exerted by the preceding strata of layers\cite{10.1007/978-3-319-24574-4_28}.

An advantageous strategy is to systematically augment the receptive field by orchestrating a series of layers equipped with small convolution kernels\cite{krizhevsky2012imagenet}. This methodology not only facilitates more proficient extraction of intricate patterns inherent in the input data but also promotes computational efficiency and reduces the propensity for overfitting through the retention of a minimalistic parameter setup in the network architecture.

The integration of non-linear activation functions in each convolutional layer further accentuates the non-linear properties of the network, thereby empowering it to adeptly navigate and adapt to more sophisticated data patterns. An exemplar in this regard is the VGG network architecture, which judiciously employs multiple strata of 3x3 convolution kernels to construct a deep network, demonstrating the effectiveness and efficiency of this strategy; it avails a richer understanding of the information encompassed in the data, eschewing the necessity for larger convolution kernels\cite{simonyan2015deep}.

In convolutional neural network paradigms, neurons exhibit a predilection for inputs within the ambit of their effective receptive field. This region, predominantly centralized and significantly more compact compared to the theoretical receptive field defined through convolutional kernels and pooling operations, elucidates a discernible trend of neural affinity towards proximal inputs over the expansive region delineated by the theoretical receptive field\cite{luo2016understanding}.

The characterization of the effective receptive field foregrounds a pivotal attribute of neuronal dynamics, portraying an augmented receptivity to stimuli originating from the central region as opposed to the diminished responsiveness elicited by peripheral inputs. This inherent central bias necessitates an accentuated focus on this critical region during the developmental and optimization stages of neural network architectures, underlining its substantial bearing on the neuron’s response kinetics\cite{10.1007/978-3-030-01252-6_24}.

Utilizing the principles underlying the effective receptive field can substantively facilitate the enhancement of network performance. Initially, it delineates the pathway for the meticulous selection of dimensions and count of convolutional layers during the network’s architectural genesis, thereby equipping the network with the prowess to decode more intricate patterns embedded in the input data\cite{li2017not}. Furthermore, it promotes a rationalized approach to network structural design, offering improved governance over the receptive field dynamics, and consequently fine-tuning the network's performance metrics.

In addition, the theoretical receptive field undergoes a progressive expansion with the deepening of the network, augmenting the network’s competence in assimilating a more extensive contextual spectrum embedded within the input data. However, it remains incumbent to recognize that notwithstanding the depth of the network, the primary influence exerted on a neuron predominantly subsists within a restricted peripheral locale, synonymous with the effective receptive field.

Initially conceived for applications in image processing and computer vision, CNNs excel in discerning patterns and idiosyncratic features within localized segments of a data set\cite{Liu2018}. While predominantly utilized in the analysis of image data, CNNs can equally be deployed in the sphere of time series analysis, offering a potent tool for the elucidation of local patterns and trends intrinsic to a series\cite{thill2021temporal}\cite{cui2016multiscale}.

In this context, one-dimensional time series data can theoretically be construed as a "one-dimensional image." Herein, convolutional strata are leveraged to pinpoint salient patterns or aberrations within the data corpus\cite{Goldberg2016}\cite{Ignatov2018}. This methodology facilitates the harnessing of local structural nuances present in the data and is adept at achieving translation invariance by identifying akin patterns manifested at divergent time intervals\cite{Goldberg2016}\cite{Hinton2012}. Moreover, the precept of weight sharing endemic to CNNs engenders a reduction in the parameter space, thereby mitigating computational exigencies and attenuating memory allocation demands\cite{wang2016time}.

Nevertheless, the employ of CNNs in time series analysis has yet to become mainstream, largely due to the intricate dynamics of sequential dependencies and temporal fluctuations that characterize time series data. Such complexities are ostensibly better navigated through the adoption of Recurrent Neural Networks (RNNs) or Long Short-Term Memory (LSTM) networks, which showcase a superior proficiency in capturing extended temporal dependencies — a faculty of critical import in myriad time series undertakings\cite{fawaz2019deep}.

\section{Recurrent Neural Network, Long Short-Term Memory network}

Multidimensional time series data, also known as multivariate time series data, represents a more complex data structure where each time stamp is associated not just with a single observation but with multiple observations or variables. We can formalize it as a function:

\[
X(t): T \rightarrow \mathbb{R}^d
\]

where:
\begin{itemize}
    \item \( X(t) \) — Time series
    \item \( T \) — Time domain
    \item \( \mathbb{R}^d \) — \( d \)-dimensional real space
    \item \( d \) — Dimensionality of the data
\end{itemize}

For a \( d \) dimensional time series, we have a \( d \) dimensional vector at each time \( t_i \):

\[
x(t_i) = [x_1(t_i), x_2(t_i), \ldots, x_d(t_i)]
\]

DNNs indeed have the capacity to accommodate and fit all time point data in a time series all at once\cite{ismailfawaz2019}\cite{bagnall2017}. This approach allows the network to capture complex patterns and dependencies at each time point. However, it also brings forth a series of issues:

\begin{enumerate}
    \item \textbf{Difficulty in Handling Long-term Dependencies:} As the information has to pass through many layers, DNNs might become less efficient in capturing dependencies in long-term time series\cite{ismailfawaz2019}.
    \item \textbf{Prone to Overfitting:} DNNs might tend to overfit, especially when attempting to learn a large number of parameters when the input data have high dimensions and complex structures.
    \item \textbf{Computational Demands:} DNNs generally require a substantial amount of computational resources to handle large-scale input data and learn numerous parameters, potentially leading to decreased efficiency and increased computational costs.
\end{enumerate}

Facing these challenges, Recurrent Neural Networks (RNNs) offer an elegant solution\cite{Elman1990}. They are specifically designed to handle sequence data, presenting the following advantages:

\begin{enumerate}
    \item \textbf{Handling Variable-length Sequences:} RNNs can process sequences step by step, allowing them to handle inputs of variable lengths.
    \item \textbf{Capturing Time Dependencies:} By maintaining an internal state to track previous inputs, RNNs can more efficiently capture time dependencies, including long-term dependencies.
\end{enumerate}

RNNs are designed to work with sequential data, representing a mapping from sequences of vectors in a multidimensional space to another multidimensional space (which could be a scalar, a vector, or even another sequence)\cite{Auli2013}\cite{Sutskever2014}. A distinguishing feature of RNNs is the use of shared weights across all time steps, facilitating the effective handling of sequences of variable lengths. Formally, we can describe the mapping as:

\[
F: \bigcup_{T=0}^{\infty} \mathbb{R}^{T \times d} \rightarrow \bigcup_{T=0}^{\infty} \mathbb{R}^{T \times p}
\]

where:
\begin{itemize}
    \item \( F \) — The mapping facilitated by the RNN
    \item \(W_{hh}\), \(W_{xh}\), and \(b_h\) — The weight matrix and bias vector that is shared across all time steps in the RNN
    \item \( T \) — The (potentially variable) length of the time series
    \item \( d \) — The number of features at each time step in the input space
    \item \( p \) — The number of features at each time step in the output space
    \item \( \bigcup_{T=0}^{\infty} \mathbb{R}^{T \times d} \) — The space of all possible input sequences, with T varying from 0 to infinity
    \item \( \bigcup_{T=0}^{\infty} \mathbb{R}^{T \times p} \) — The space of all possible output sequences, which can range from a scalar (when \( p = 1 \) and \( T = 1 \)) to a multi-dimensional time series (when \( p > 1 \) and \( T > 1 \))
\end{itemize}

At each time step \( t \), the RNN receives an input vector from \( \mathbb{R}^d \) and utilizes the shared weight matrix \( W \) to process inputs over \( T \) time steps and generate an output in \( \mathbb{R}^p \). The variability in the length of the time series \( T \) enables the RNN to handle sequences of different lengths efficiently.

The fundamental working principle of an RNN is to receive an input at each time step and update its hidden state, which contains information about the sequence so far:

\[
h_t = \sigma(W_{hh} h_{t-1} + W_{xh} x_t + b_h)
\]

In this equation, \(W_{hh}\), \(W_{xh}\), and \(b_h\) are the learnable parameters of the network, optimized through the training process.

While deep neural networks can receive the entire time series as input to capture complex patterns, they struggle with handling long-term dependencies and avoiding overfitting. In contrast, recurrent neural networks effectively address these issues by processing sequences step by step and maintaining an internal “memory” to efficiently track previous inputs, making them a powerful tool for handling time series data. Nevertheless, RNNs have exhibited limitations in learning and retaining extended sequential dependencies, primarily due to issues such as gradient vanishing and gradient exploding\cite{KolenKremer2001}. To overcome these hurdles, LSTM networks were devised to capture long-term dependencies in time series through a unique “cell state” and “gate” mechanisms\cite{hochreiter1997long}\cite{Greff2015LSTMAS}.

In LSTMs, each cell contains the following components to control the flow of information:

\begin{itemize}
    \item \textbf{Cell State (\( C_t \))}: This is the core component of the LSTM cell, responsible for maintaining and transmitting information throughout the sequence. The cell state is updated as follows:
      \[
      C_t = f_t \cdot C_{t-1} + i_t \cdot \text{tanh}(W_C \cdot [h_{t-1}, x_t] + b_C)
      \]
      where:
      \begin{itemize}
          \item \( C_t \) and \( C_{t-1} \) are the cell states at the current and previous time steps, respectively.
          \item \( f_t \) is the activation of the forget gate, controlling how much of the past information we want to forget.
          \item \( i_t \) is the activation of the input gate, controlling how much new information we want to add.
          \item \( W_C \) and \( b_C \) are the weight matrix and bias term learned through training.
      \end{itemize}

    \item \textbf{Input Gate (\( i_t \))}: Determines how much new information to add to the cell state. It is calculated using a sigmoid function that compresses the value between 0 and 1, as follows:
      \[
      i_t = \sigma(W_i \cdot [h_{t-1}, x_t] + b_i)
      \]

    \item \textbf{Forget Gate (\( f_t \))}: Determines how much of the old information to retain. It is controlled through a sigmoid function, as below:
      \[
      f_t = \sigma(W_f \cdot [h_{t-1}, x_t] + b_f)
      \]

    \item \textbf{Output Gate (\( o_t \))}: Controls how much information from the cell state to output to the next layer in the network. It uses a sigmoid function to calculate the quantity of the output, as:
      \[
      o_t = \sigma(W_o \cdot [h_{t-1}, x_t] + b_o)
      \]
      \[
      h_t = o_t \cdot \text{tanh}(C_t)
      \]
\end{itemize}

At the core is the additive update strategy for the cell state, facilitating more appropriate gradient propagation\cite{Jzefowicz2015AnEE}. The gated units can determine how much gradient information to forget at different time points, and the values for these gated units are learned through the joint influence of the hidden state and the current input\cite{Guo2021MALSTMAM}. Therefore, even though the weight matrices for the gated units remain the same across all time points, they can adopt different values at distinct times to control the flow of information and the propagation of gradients.

In LSTMs, both the cell state and the hidden state can be passed to the next time step as inputs to the network, helping alleviate the vanishing gradient issue and ensuring that the network can learn long-term dependencies more effectively. This is achieved through a gating mechanism that allows the network to learn how to control the flow of information more adeptly, and to adjust the gradients appropriately to avoid vanishing or exploding\cite{Yu2019ARO}.

Thus, LSTMs can not only handle sequence data of varying lengths more effectively but also better capture long-term dependencies in time-series data through its special structure, making it a very powerful and flexible tool for a wide range of sequence-to-sequence learning tasks.

\chapter{Path Develepment}
\label{chapterlabel3}
\section{Lie Group}

A Lie group is a group \( G \) that is also a differentiable manifold, such that the group operations of multiplication and inversion are smooth maps.

Lie algebra is associated with a Lie group and can be understood through the tangent space to the Lie group at the identity element. Given the vector space \(V\), we define a bilinear operation named "Lie bracket," \([ \cdot, \cdot ]: V \times V \rightarrow V\), satisfying the following properties:

\begin{enumerate}
\item \textbf{Bilinearity}: 
\[
[X, Y + Z] = [X, Y] + [X, Z], \quad [X + Y, Z] = [X, Z] + [Y, Z].
\]
\item \textbf{Anticommutativity}: 
\[
[X, Y] = - [Y, X].
\]
\item \textbf{Jacobi identity}: 
\[
[X, [Y, Z]] + [Y, [Z, X]] + [Z, [X, Y]] = 0.
\]
\end{enumerate}

Given a Lie group \(G\), we can define its associated Lie algebra through the tangent space at the identity element of the group, denoted as \(e\). The Lie algebra, often denoted as \(\mathfrak{g}\), is defined as the tangent space at \(e\), i.e., 
\[
\mathfrak{g} = T_eG.
\]
In specific contexts, the Lie bracket can take various concrete forms. For instance, in the matrix Lie algebras, it is often defined as the commutator of matrices:

\[
[X, Y] = XY - YX.
\]

Here, \(X\) and \(Y\) are matrices, and the multiplication is the usual matrix multiplication.

\textbf{Definition (Exponential Map)}:

The exponential map is a crucial tool in the context of Lie algebras and Lie groups. For a given Lie algebra \(\mathfrak{g}\), the associated exponential map is a map from \(\mathfrak{g}\) to its associated Lie group \(G\), defined as follows:

\[
\exp : \mathfrak{g} \to G,
\]

The mapping is defined by the following series:

\[
\exp(X) = \sum_{n=0}^{\infty} \frac{1}{n!} X^n,
\]

where \(X\) is an element of \(\mathfrak{g}\), and \(X^n\) refers to the \(n\)-th Lie product of \(X\) with itself (in the matrix representation of the Lie algebra, it equates to the ordinary matrix power)\cite{hall2000elementary}.

This map satisfies several important properties:

\begin{enumerate}
\item \(\exp(0) = e\), where \(0\) is the zero element of the Lie algebra \(\mathfrak{g}\), and \(e\) is the identity element of the Lie group \(G\).

\item The exponential map is a local diffeomorphism, which means that it establishes a one-to-one mapping between a small neighborhood around \(0\) in \(\mathfrak{g}\) and a neighborhood around \(e\) in \(G\).

\item Under certain conditions (for instance when the Lie algebra is a Connected and Compact Lie algebra), the exponential map is surjective.
\end{enumerate}

Defining an exponential map does not automatically yield a Lie group. Actually, the exponential map is a map from a given Lie algebra to a known Lie group.

For clarity, let's consider a straightforward example: \(GL(n, \mathbb{R})\) is the group of all \(n \times n\) invertible real matrices, which forms a Lie group. Its corresponding Lie algebra is the set of all \(n \times n\) real matrices, denoted as \(\mathfrak{gl}(n, \mathbb{R})\). For any element \(X\) in this Lie algebra, the exponential map is defined as follows:

\[
\exp(X) = \sum_{n=0}^{\infty} \frac{1}{n!} X^n,
\]

where \(X^n\) is the standard matrix power (i.e., the \(n\)-th product of \(X\) with itself). This exponential map is a smooth map from \(\mathfrak{gl}(n, \mathbb{R})\) to \(GL(n, \mathbb{R})\)\cite{hall2000elementary}.

Here are the general definitions of some common Lie groups:
\begin{align*}
\text{GL}(n, \mathbb{R}) & := \{A \in \mathbb{R}^{n\times n} \,|\, \text{det}(A) \neq 0\}, \\
\text{SL}(n, \mathbb{R}) & := \{A \in \mathbb{R}^{n\times n} \,|\, \text{det}(A) = 1\}, \\
\text{O}(n) & := \{A \in \mathbb{R}^{n\times n} \,|\, A^{\top}A = I_n\}, \\
\text{SO}(n) & := \{A \in \mathbb{R}^{n\times n} \,|\, A^{\top}A = I_n, \, \text{det}(A) = 1\}, \\
\text{U}(1) & := \{z \in \mathbb{C} \,|\, |z| = 1\}, \\
\text{SU}(n) & := \{A \in \mathbb{C}^{n\times n} \,|\, A^*A = I_n, \, \text{det}(A) = 1\}, \\
\text{Sp}(2m, \mathbb{R}) & := \{A \in \text{gl}(2m; \mathbb{R}) \,|\, A^{\top}J_m + J_m A = 0\},
\end{align*}

where \(J_m\) \text{ is a fixed } \(2m \times 2m\) \text{ symplectic matrix.}

\section{Optimization of Matrix Lie Groups}
In the context of manifold optimization, the problem is typically formulated as seeking a point \( x \in \mathcal{M} \) where \( \mathcal{M} \) denotes a manifold, that minimizes (or maximizes) a designated function \( f: \mathcal{M} \rightarrow \mathbb{R} \).

Formally, the problem can be depicted as:

\[
\text{{minimize}} \quad f(x) \quad \text{{subject to}} \quad x \in \mathcal{M}
\]

Where:

\begin{itemize}
\item \( f: \mathcal{M} \rightarrow \mathbb{R} \) is the objective function we intend to minimize.
\item \( \mathcal{M} \) represents the manifold constituting the constraint set harboring all feasible solutions.
\end{itemize}

Defining an optimization problem on a manifold without additional structure can be very challenging because we lack a natural way to define "distance" or "length", making it impractical or at least very difficult to define methods such as gradient descent, the Riemannian metric allows us to define concepts such as "distance" and "angle," enabling us to define and solve optimization problems more naturally, for instance, through gradient descent methods\cite{Bonnabel2011StochasticGD}\cite{Bcigneul2018RiemannianAO}.

Consider a matrix Lie group, which also constitutes a Riemannian manifold. In this context, we define a differentiable mapping \(\phi: \mathcal{M} \to \mathbb{R}^n\). Our objective is to optimize a function \(f: \mathbb{R}^n \to \mathbb{R}\) composed as \(f(\phi(x))\), where \(x\) resides in the manifold \(\mathcal{M}\).

\begin{theorem}\cite{10.5555/3454287.3455108}\cite{Hunacek2008}
\(f\) is a mapping,
\(
f : gl(n; \mathbb{C}) = \mathbb{C}^{n \times n} \to \mathbb{R}
\), and \(exp\) represent the matrix exponential, \(A\) is an arbitrary matrix in \(gl(n; \mathbb{C})\). \text{ We have the following}.
\[
\nabla( f \circ {exp})(A) = (d {exp})_{A^\top} (\nabla f ({exp}(A)))
\]
\[
(d\exp)_A(X) = \sum_{k=0}^{\infty} \frac{(-ad(A))^k}{(k+1)!} (\exp(A)X)
\]
\[
\text{where }\text{ad}\text{ is the adjoint action defined by }\text{ad}(X)Y := XY - YX.
\]

\end{theorem}
Typically, optimization problems involve trying to find an optimal solution of a function over a specific manifold. A complex non-linear optimization problem can be simplified to a more manageable problem in Euclidean space. By utilizing a "trivialization" mapping, a subset of the Lie group can be mapped to \( \mathbb{R}^n \). In this newly defined space, the optimization process is more straightforward\cite{10.5555/3454287.3455108}\cite{Casado2020}. This strategy can be expressed mathematically as

\begin{equation}
\begin{array}{l}
\text{{minimize}} \quad f(x) \\
\quad x \in \mathcal{M}
\end{array}
\quad \xrightarrow{} \quad 
\begin{array}{l}
\text{{minimize}} \quad f(\Phi(a)) \\
\quad a \in \mathbb{R}^n
\end{array}
\end{equation}

\section{Path Development Layer}

The path development layer and path signatures share some similarity as both are driven by time series to some extent\cite{davie2007differential}\cite{chevyrev2016primer}.

\textbf{Definition (Path development)}\cite{lou2022path}:

Let \(G\) be a finite-dimensional Lie group with Lie algebra \(\mathfrak{g}\). Let \(M : \mathbb{R}^d \rightarrow \mathfrak{g} \subset \text{gl}(m; F)\) be a linear map and let \(X \in \mathcal{V}^1([0, T]; \mathbb{R}^d)\) be a path, where \(\mathcal{V}^1\) represents the space of absolutely continuous functions. The path development (a.k.a. the Cartan development) of \(X\) on \(G\) under \(M\) is the solution to the equation

\[
dY_t = Y_t \cdot M(dX_t) \quad \text{for all} \quad t \in [0, T], \quad Y_0 = e,
\]

where \(Y_t\) represents the path development at time \(t\), and \(e\) is the identity element of the Lie group \(G\). Here the multiplication is matrix multiplication.

For a linear path \(X \in \mathcal{V}^1([0, T]; \mathbb{R}^d)\), its development on a matrix Lie group \(G\) under \(M \in L(\mathbb{R}^d, \mathfrak{g})\) is

\[
D_M(X)_{0,t} = \exp\left(M(X_t - X_0)\right)\cite{lou2022path}.
\]

This approach can be seen as employing the Picard iteration method based on the compression mapping principle in mathematical terms.

The characteristics of the path development layer resemble those of path signatures, as they both exhibit time Invariance and Multiplicative property\cite{lou2022path}. The path development layer does not record the occurrence speed of time-series data. When analyzing content where the speed is determined during recording, such as in stock data and electrocardiograms, it can remove the irrelevant parts. 

\begin{theorem}(Invariance under time-reparametrisation).
Let \(X \in \mathcal{V}^1([0, T]; \mathbb{R}^d)\) and let \(\lambda\) be a non-decreasing function from \([0,T_1]\) onto \([0,T_2]\). Define \(X_t^\lambda := X_{\lambda_t}\) for \(t \in [0,T]\). Then, for all \(M \in L(\mathbb{R}^d, \mathfrak{g})\) and for every \(u,t \in [0,T]\),

\[
D_M(X_{\lambda_u,\lambda_t}) = D_M(X^\lambda_{u,t}).
\]
\end{theorem}

Path signatures represent all the information of solutions to linear differential equations driven by paths\cite{boedihardjo2016signature}\cite{chen1958integration}. While a single path cannot be uniquely determined solely by its path signature, path signatures preserve all other information except for speed and the ability to differentiate between time-reversed paths\cite{hambly2010uniqueness}\cite{chevyrev2016primer}. Furthermore, the path development layer does not discard any other information compared to path signatures. So, the path development layer can be used for some feature extraction tasks\cite{lou2022path}.

\textbf{Definition (Path development layer)}\cite{lou2022path}:

Fix a matrix group \(G\) with Lie algebra \(\mathfrak{g}\).given \( \theta = (\theta_1, \ldots, \theta_d) \in \mathfrak{g}^d \), define

\[
M_{\theta} : \mathbb{R}^{d} \ni x = (x_1, \ldots, x_d) \mapsto \sum_{j=1}^{d} \theta_j x_j \in \mathfrak{g}.
\]

The path development layer is defined as a map 
\[
D_{\theta} : \mathbb{R}^{d \times (N+1)} \to G^{N+1} \text{ [or \(G\), resp.]} : x = (x_1, \ldots, x_N) \mapsto z = (z_1, \ldots, z_N) \text{ [or \(z_N\), resp.]}
\]

such that for each \(n \in \{0, \ldots, N - 1\}\),

\[
z_{n+1} = z_n \exp\left(M_\theta (x_{n+1} - x_n)\right),
\]

with the initial condition
\[
z_0 = Id_m.
\]

Here, \(\exp\) is the matrix exponential, \(G^{N+1}\) is the \((N+1)\)-fold Cartesian product of \(G\), and \(\theta \in \mathfrak{g}^d\) are trainable model weights, d represents the dimensionality of the channels in the input time series.

So, the forward propagation algorithm can be obtained\cite{lou2022path}.

\begin{algorithm}
\caption{forward propagation algorithm of the path development layer}
\begin{algorithmic}[1]
\State Input: \(\theta \in \mathfrak{g}^d \subset \mathbb{R}^{m \times m \times d} \text{ (Model parameters)}\)
\State \(x = (x_0, \ldots, x_N) \in \mathbb{R}^{d \times (N+1)} \text{ (input time series)}\)
\State \(m \in \mathbb{N} \text{ (order of the matrix Lie algebra)}.\)
\State \((d, N) \text{ are the feature and time dimensions of } x \text{, respectively.}\)
\State \(z_0 \leftarrow \text{Id}_{m}\)
\For{\(n \in \{1, \ldots, N\}\)}
\State \(z_n \leftarrow z_{n-1} \exp\left(M_\theta(x_n - x_{n-1})\right)\)
\EndFor
\State \(\text{Output: } z = (z_0, \ldots, z_N) \in G^{N+1} \subset \mathbb{R}^{m \times m \times (N+1)} \text{ (sequential output) or } z_N \in G \subset \mathbb{R}^{m \times m} \text{ (static output).}\)
\end{algorithmic}
\end{algorithm}

It exhibits certain similarities with RNNs, as the parameters remain constant across all time steps in this context.

\section{Backpropagation Algorithm in the Path Development Layer}

Neural network optimization is typically done using the backpropagation algorithm. The preceding theorems can help in understanding the update strategy for these parameters, leading to a backward propagation algorithm somewhat akin to recurrent neural networks.

Hang Lou, Siran Li, and Hao Ni collaborated on a paper where they discussed the backpropagation in the path development layer and provided a proof of an optimization theorem\cite{lou2022path}.

Using this theorem, we can derive the backward propagation algorithm for training the path development layer\cite{lou2022path}.

\begin{theorem}\cite{lou2022path}.
Let \(x \in \mathbb{R}^{d \times (N+1)}\), \(z = (z_0, \ldots, z_N) \in {G}^{N+1}\), \(\psi : {G} \to \mathbb{R}\), \(\theta_n \in \mathfrak{g}^d\), and \(Z_n : \mathfrak{g}^d \to {G}\) be as above. Denote by \(d_{HS} := \text{tr}(A^{\top}B)\) as the metric associated to the Hilbert--Schmidt norm on the matrix Lie algebra \( \mathfrak{g} \). For each \( \theta \in \mathfrak{g}^d \), the gradient \( \nabla_{\theta} (\psi \circ Z) \in T_{\theta}\mathfrak{g}^d \cong \mathfrak{g}^d \) (with respect to a metric arising naturally from \( d_{HS} \)) satisfies the following:

\[
\nabla_{\theta} (\psi \circ Z) = \sum_{n=1}^{N} \nabla_{\theta_n} (\tilde{\psi}_n \circ Z_n)|_{\theta_n=\theta}
\]

\[
\nabla_{\theta_n} (\tilde{\psi}_n \circ Z_n)=d_{\mathfrak{J}_n\theta_n^{\top}}\exp(\nabla_{z_n} \tilde{\psi}_n \cdot z_{n-1}) \otimes \Delta x_n.
\]

Here we denote \(\mathfrak{J}_n(\theta_n) := M_{\theta_n} (\Delta x_n)\). Also, as before, \(\Delta x_n := x_n - x_{n-1}\) and \(\cdot\) is the matrix multiplication.

\[
S_i : G \to G, \quad S_i(z_{i-1}) := z_i \quad \text{for } i \in \{1, 2, \ldots, N\}.
\]

Define \(\tilde{\psi}_n : G \to \mathbb{R}\) for \(n \in \{0, 1, \ldots, N-1\}\)
\[
\tilde{\psi}_n(z_n) := \psi\left(z_0, \ldots, z_n, S_{n+1}(z_n), S_{n+2} \circ S_{n+1}(z_n), \ldots, S_N \circ \ldots \circ S_{n+1}(z_n)\right).
\]
\end{theorem}

\begin{algorithm}
\caption{backward propagation algorithm for the path development layer}
\begin{algorithmic}[1]
\State Input: \(x = (x_0, \ldots, x_N) \in \mathbb{R}^{(N+1) \times d} \text{ (input time series)}, z = (z_0, \ldots, z_N) \in G^{N+1} \text{ (output series by the forward pass)}, \theta = (\theta_1, \ldots, \theta_d) \in \mathfrak{g}^d \subset \mathbb{R}^{m \times m \times d} \text{ (model parameters)}, \eta \in \mathbb{R} \text{ (learning rate)}, \hat{\psi} : \text{gl}(m, \mathbb{C})^{N+1} \to \mathbb{R} \text{ (loss function)}\)
\State \(\text{Initialize } \omega \leftarrow 0, a \leftarrow 0, \mathfrak{J} \leftarrow 0.\)
\For{\(n \in \{N, \ldots, 1\}\)}
\State \(\text{Compute } a \leftarrow \left. {\partial_{y_n} \hat{\psi}} \right|_{y=z} + a \cdot \mathfrak{J}\)
\State \(\text{Compute } \mathfrak{J} \leftarrow M_{\theta}(\Delta x_n).\)
\State \(\text{Compute } \omega \leftarrow \omega + d_{\mathfrak{J}^{\top}} \exp(a \cdot z_{n-1}) \otimes \Delta x_n.\)
\EndFor
\State \(\theta \leftarrow \theta - \eta \cdot \omega.\)
\For{\(i \in \{1, \ldots, d\}\)}
\State \(\theta_i \leftarrow \text{Proj}(\theta_i) \in \mathfrak{g}.\) \text{Proj} is a projection: \(\operatorname{gl}(m, \mathbb{C}) \rightarrow \mathfrak{g}\)
\EndFor
\State Output: return \(\theta\) (updated model parameter)
\end{algorithmic}
\end{algorithm}


\chapter{ECG Data Analysis}
\label{chapterlabel4}
\section{Models for ECG}
\subsection{ECG Dataset}

The electrocardiogram data was recorded in the Cardiology Department of Jiangsu Provincial People's Hospital in China. Each sample consists of 12-channel data with 10,000 time points. It includes data from 871 atrial fibrillation patients, capturing their daily electrocardiograms and continuous monitoring of their subsequent lives, there are 143 positive samples and 622 negative samples. A positive result indicates the occurrence of a stroke in the following period, while a negative result means no stroke occurred.

What sets this paper apart from others is that, while most papers typically focus on detecting whether an ECGs is atrial fibrillation (AFib) or normal, this paper is predicting the probability of stroke in AFib patients based on their routine ECGs. In this case, the positive samples include both AFib and normal ECGs, while the negative samples include both AFib and normal ECGs. Furthermore, both positive and negative samples may contain instances of premature ventricular contractions (PVCs), premature atrial contractions (PACs), ventricular tachycardia (VT), supraventricular tachycardia (SVT), and normal sinus rhythm.

A typical positive ECG record and a negative ECG record look like Figure 4.1 and Figure 4.2.

\begin{figure}[h] 
  \centering
  \includegraphics[width=1\textwidth]{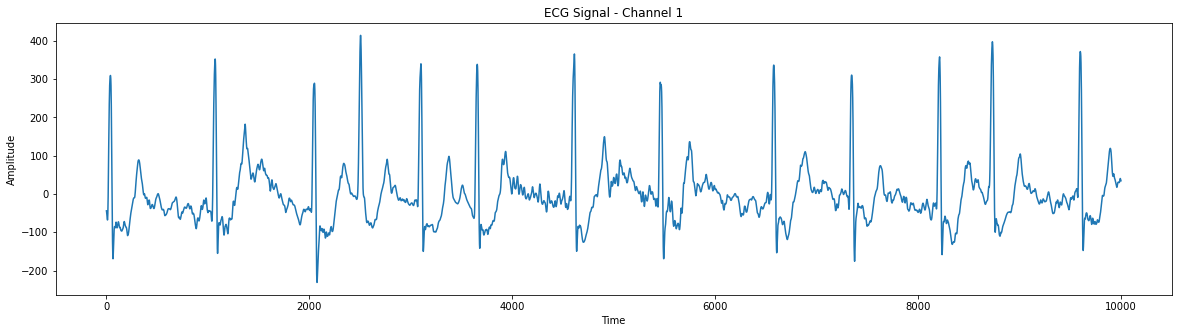} 
  \caption{positive ECG}
  \label{fig:example} 
\end{figure}

\begin{figure}[h] 
  \centering
  \includegraphics[width=1\textwidth]{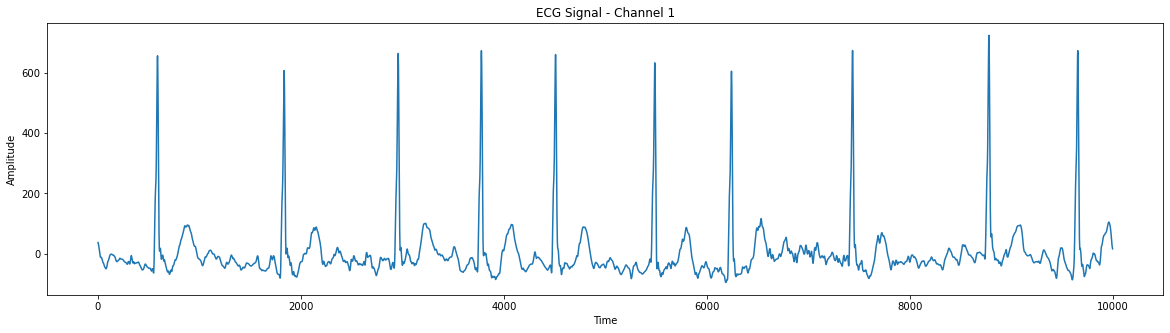} 
  \caption{negative ECG}
  \label{fig:example} 
\end{figure}

\subsection{Data Preprocessing and Wavelet Transformation}

Most ECG recordings inherently introduce some level of noise, which can lead to overfitting, especially when the dataset is limited. Therefore, one common strategy in machine learning for ECG analysis is to use wavelet transformation to reduce noise in the recorded data\cite{kumar2021stationary}. This helps mitigate overfitting and is a prevalent approach in handling ECG data. Using the Daubechies 6 (db6) wavelet function and level 4 for wavelet transformation is a common parameter configuration when processing ECG data.

The changes in the ECG before and after wavelet transformation resemble these two graphs Figure 4.3 and Figure 4.4.

\begin{figure}[h] 
  \centering
  \includegraphics[width=1\textwidth]{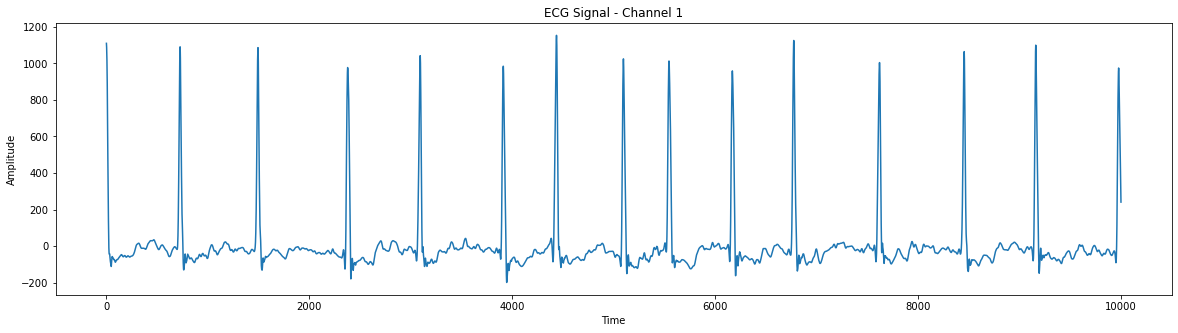} 
  \caption{Before wavelet transformation processing}
  \label{fig:example} 
\end{figure}

\begin{figure}[h] 
  \centering
  \includegraphics[width=1\textwidth]{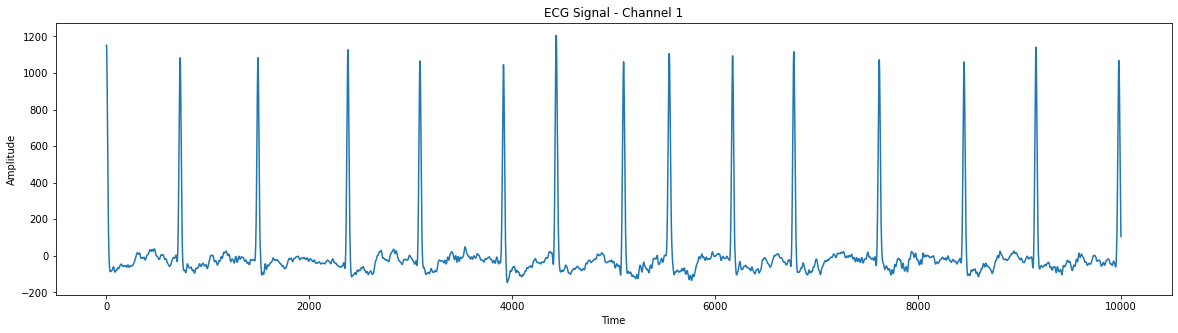} 
  \caption{After wavelet transformation processing}
  \label{fig:example} 
\end{figure}

In the model, each ECG record is input into the same wavelet transformation for processing, followed by feeding it into a neural network.

\subsection{Data Augmentation}

The positive samples in the data are slightly less than the negative samples, so using conventional machine learning strategies may lead the model to classify all samples as negative. Some researchers have employed alternative loss functions, but data augmentation is also a viable approach\cite{garcia2007empirical}\cite{lin2020focal}. In this paper, the SMOTE data augmentation algorithm was utilized\cite{chawla2002smote}. Furthermore, since there is no medical research on the features of atrial fibrillation leading to stroke, employing alternative data augmentation strategies might result in some class misclassification errors.

The working principle of SMOTE is to create synthetic samples for the minority class rather than simply duplicating minority class samples. It achieves this by randomly selecting a minority class sample and one of its k nearest neighbors. It then uses the linear equation between these two points to generate a new synthetic point\cite{chawla2002smote}.

Specifically, given a minority class sample "a" and one of its nearest neighbors "b," a new synthetic sample "x" would be created satisfying the following condition:

\[
x = a + \lambda \cdot (b - a)
\]

where \(\lambda\) is a random number between 0 and 1.

In this way, the SMOTE algorithm creates a more enriched and diverse training dataset, which can help improve the performance of the classifier, especially regarding the classification performance of the minority class.

\subsection{Cross-Entropy Loss Function}

This paper uses the cross-entropy loss function as the loss function, employs \(L^2\) regularization to address overfitting issues due to limited data, with a regularization weight of \texttt{L2\_Weight}. Cross-entropy is a function commonly used to measure the difference between two probability distributions. In machine learning and deep learning, cross-entropy is often used as a loss function for classification problems, especially when training neural networks\cite{shalev2014model}.

The cross-entropy function is defined as follows:

\[
H(p, q) = - \sum_x p(x) \log q(x)
\]

Where:
\begin{itemize}
    \item \( H(p, q) \) is the cross-entropy
    \item \( p(x) \) is the true probability distribution
    \item \( q(x) \) is the predicted probability distribution
    \item The formula sums over all possible events \( x \)
\end{itemize}

In binary classification problems, the cross-entropy loss can be simplified to:

\[
H(p, q) = - [y \log(\hat{y}) + (1 - y) \log(1 - \hat{y})]
\]

Where:
\begin{itemize}
    \item \( y \) is the true label, which can be 0 or 1
    \item \( \hat{y} \) is the predicted probability, representing the predicted probability of the label being 1
\end{itemize}
\subsection{Network Architecture}

In this paper, we employ a method that integrates Convolutional Neural Network (CNN), Long Short-Term Memory (LSTM) network, and Path Development Layer. Before being fed into the network, the data first undergoes wavelet transformation. The data initially passes through a CNN with a specified number of layers, denoted as \texttt{CNN\_LayerNumber}. Following each convolutional layer, there is a pooling layer employing max-pooling. To streamline the hyperparameter tuning, the same values for the number of filters and strides are used in each layer, noted as \texttt{CNN\_kernel} and \texttt{CNN\_stride}, and the number of channels in each convolutional layer remains consistent, denoted as \texttt{CNN\_Channel}. After each convolutional layer, a BatchNorm layer is applied \cite{ioffe2015batch}. Subsequently, the data is input into an LSTM network with a specified number of layers \texttt{LSTM\_LayerNumber} and a defined number of hidden nodes \texttt{LSTM\_Number}. The output from the LSTM represents an intermediate time sequence.

In this manner, by incorporating a composite model of Convolutional Neural Network, Long Short-Term Memory network, and Path Development Layer, this paper effectively captures the spatiotemporal characteristics of the data while preprocessing the input data and adjusting network parameters, providing new insights and methods for the optimization and application of deep learning models.

This intermediate time sequence is then input into the Path Development Layer, where orthogonal matrices serve as the internal Lie group, and the matrix length is set as \texttt{DEV\_Number}. The final matrix in the generated matrix sequence is selected as the result. Following this, the data passes through a neural network with a length of \texttt{DNN\_Number}. The output, represented as a 2-dimensional array, goes through a softmax layer, which represents the predicted probabilities of positive and negative classes. It is crucial to emphasize that, in the context of time series data, a matrix is generated for each channel within the path development layer. Consequently, when dealing with a high number of channels, the resulting matrix output can become significantly voluminous. This characteristic stands as a notable point of differentiation between the path development layer and the long short-term memory network.

During training, neural networks are typically deep, so the activation function used in the models in this paper is ReLU.

The \textbf{ReLU} (Rectified Linear Unit) activation function is a pivotal element in deep learning architectures, prominently utilized owing to its favorable properties in facilitating the learning process in neural networks\cite{10.5555/3104322.3104425}. The function is mathematically delineated as:

\[
f(x) = \max(0, x)
\]

The ReLU function extends several substantial benefits, which are delineated below:

\begin{enumerate}
    \item \textbf{Induction of Non-linearity}: Albeit it appears linear, ReLU imparts non-linearity due to its segmented linear behavior, an essential trait that equips neural networks with the ability to learn from the error backpropagated and to approximate intricate and non-linear mappings from inputs to outputs.
    \item \textbf{Computational Efficiency}: The simplicity in the mathematical formulation of ReLU makes it computationally less intensive compared to its counterparts such as sigmoid or tanh, thereby expediting the network training process\cite{10.5555/3104322.3104425}.
    \item \textbf{Mitigating the Vanishing Gradient Problem}: ReLU effectively mitigates the gradient problem encountered in deep neural networks. This is predominantly because it maintains a constant gradient of one for all positive inputs, facilitating gradient descent optimization to update the weights even in deeper layers without the gradients tending to vanish, a phenomenon quite prevalent with other activation functions\cite{10.5555/3104322.3104425}.
    \item \textbf{Promotion of Sparse Representations}: ReLU encourages sparse representations, which tend to render the network robust and help in learning more nuanced features, enhancing both the training and generalization phases\cite{7410480}\cite{pmlr-v15-glorot11a}.
\end{enumerate}

Notwithstanding its merits, the ReLU function harbors certain demerits:

\begin{enumerate}
    \item \textbf{Dying ReLU Phenomenon}: During the training phase, neurons can sometimes saturate, yielding zero gradients for negative inputs. This induces a permanent learning cessation for those neurons, a scenario colloquially termed as the "dying ReLU" problem\cite{clevert2016fast}.
    \item \textbf{Non-symmetric Behavior}: ReLU has a non-zero mean, which can introduce bias in subsequent layers. The higher the correlation, the greater the bias\cite{clevert2016fast}.
\end{enumerate}

The network architecture can be understood using Figure 4.5 and Figure 4.6. In the actual model, multiple convolutional layers are included.

\begin{figure}[h] 
  \centering
  \includegraphics[width=1\textwidth]{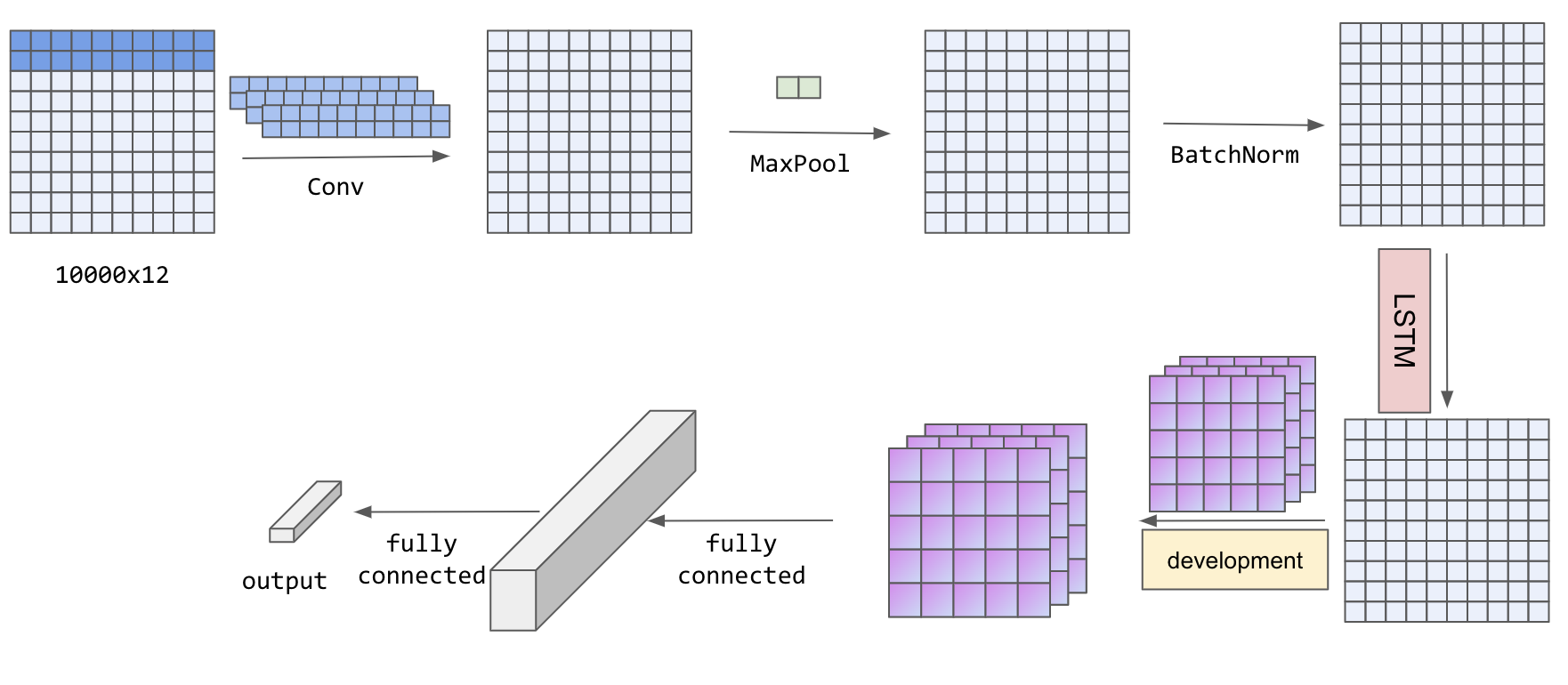} 
  \caption{Network architecture diagram}
  \label{fig:example} 
\end{figure}

\begin{figure}[h] 
  \centering
  \includegraphics[width=1\textwidth]{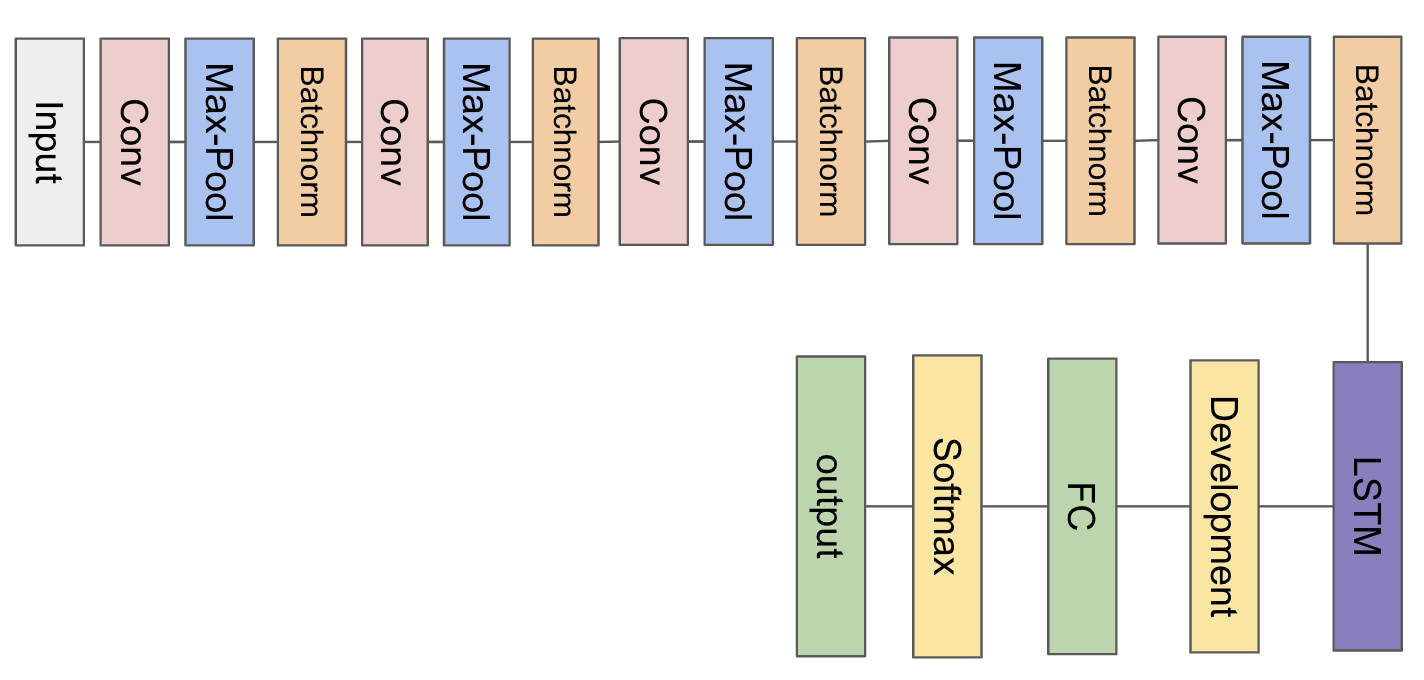} 
  \caption{Network architecture diagram}
  \label{fig:example} 
\end{figure}

To facilitate a comprehensive and stringent comparison, an initial model, designated as \texttt{LSTM\_Model}, which is devoid of the Path Development Layer, is employed as a foundational baseline. This model maintains consistency in all other aspects, with the sole distinction being the absence of the Path Development Layer. Detailed insights into the architecture of this network can be gleaned from Figure 4.7 and Figure 4.8.

With the aim of conducting a direct comparative analysis between the learning capacities of LSTM networks and Path Development Layers, the LSTM layer within the \texttt{LSTM\_Model} is replaced with a Path Development Layer, giving rise to a novel model, \texttt{DEV\_Model}. The intricacies of the architecture of this newly-formed network can be further elucidated through Figure 4.9 and Figure 4.10.

In addition, reference is made to a previously described comprehensive model that incorporates both the LSTM layer and the Path Development Layer. Subsequently, a series of training procedures are initiated to assess and compare the performance characteristics of these three models, namely, the \texttt{LSTM\_Model}, the \texttt{DEV\_Model}, and the aforementioned comprehensive model encompassing all layers.

\begin{figure}[h] 
  \centering
  \includegraphics[width=1\textwidth]{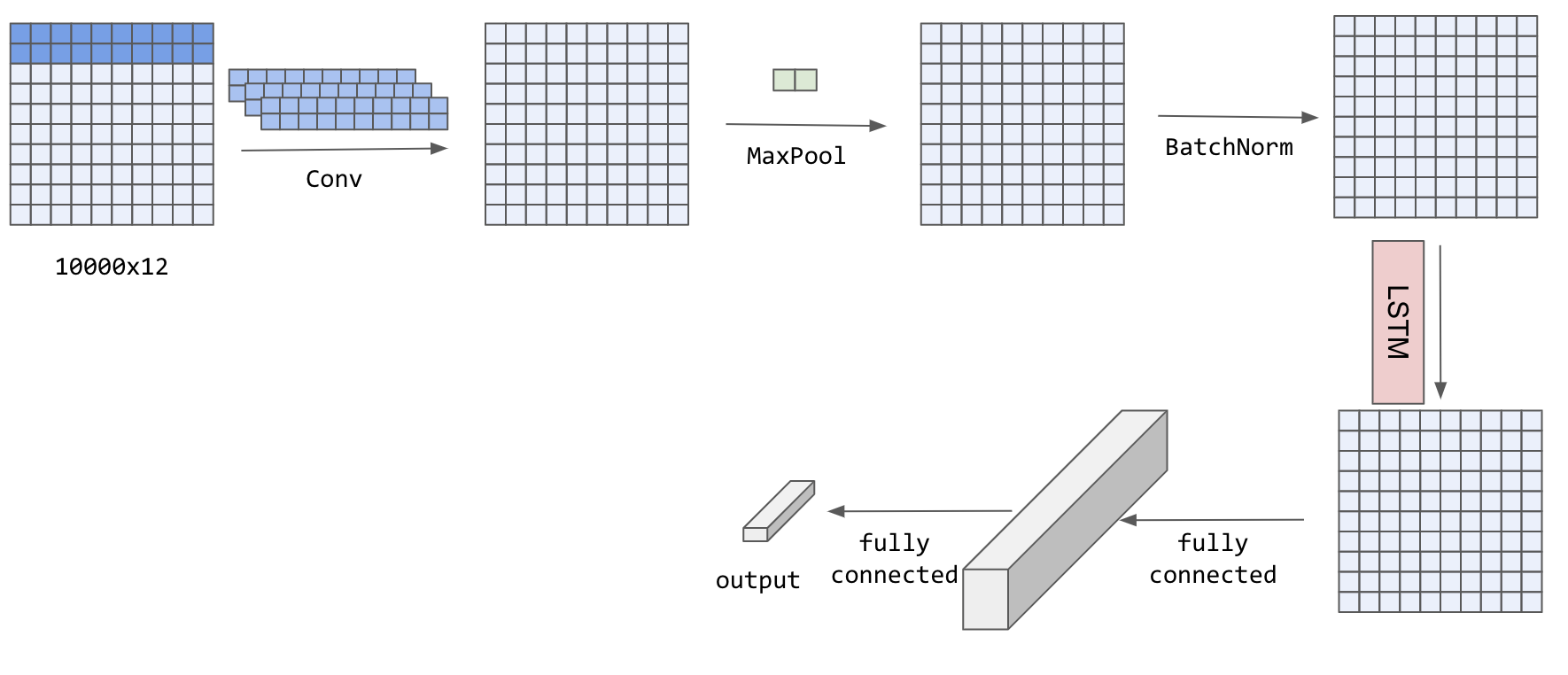} 
  \caption{LSTM\_Model}
  \label{fig:example} 
\end{figure}

\begin{figure}[h] 
  \centering
  \includegraphics[width=1\textwidth]{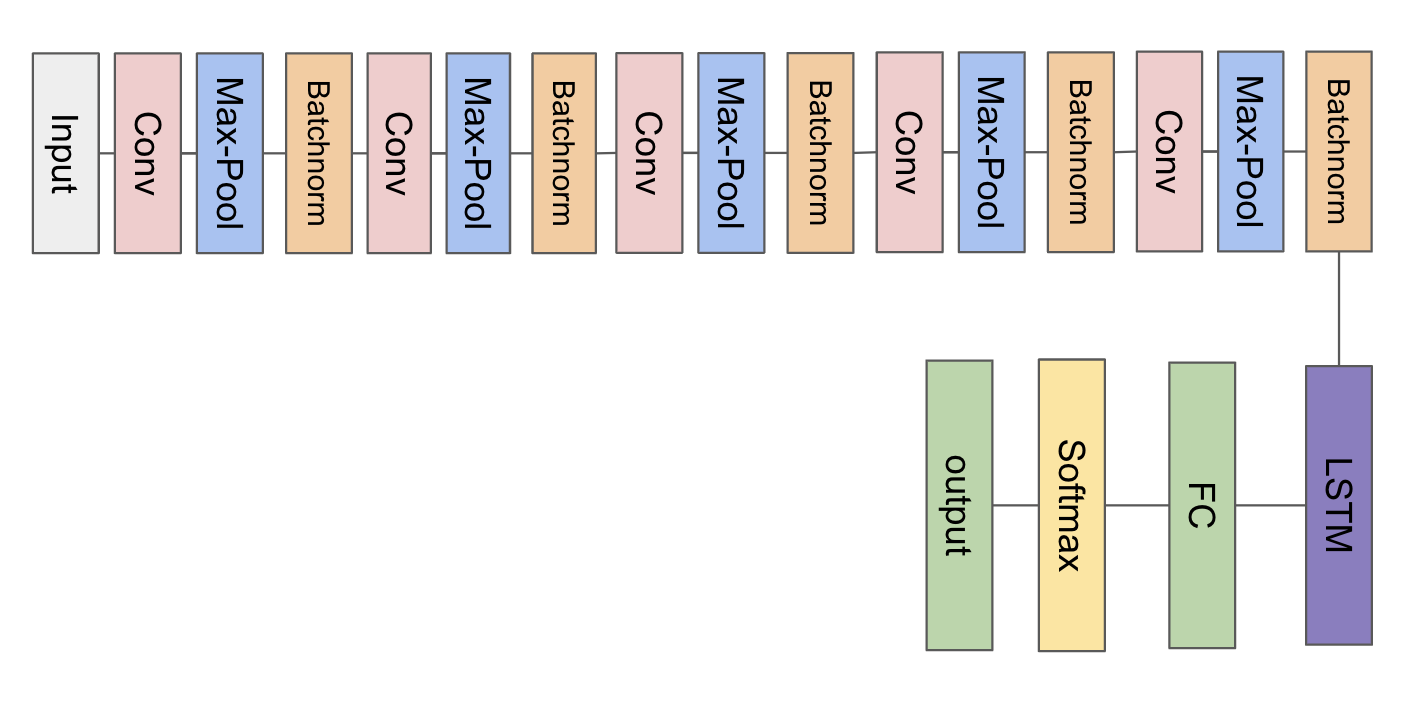} 
  \caption{LSTM\_Model}
  \label{fig:example} 
\end{figure}

\begin{figure}[h] 
  \centering
  \includegraphics[width=1\textwidth]{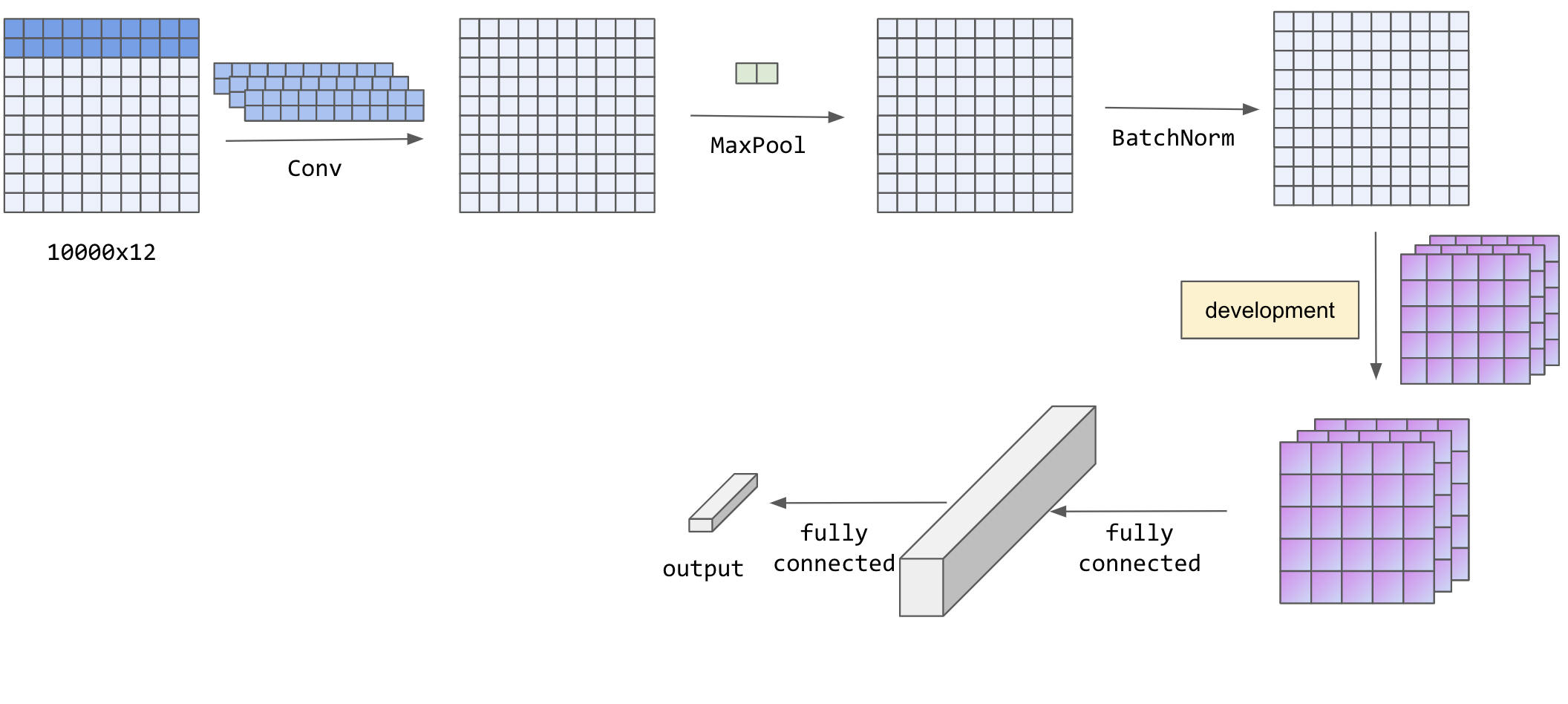} 
  \caption{DEV\_Model}
  \label{fig:example} 
\end{figure}

\begin{figure}[h] 
  \centering
  \includegraphics[width=1\textwidth]{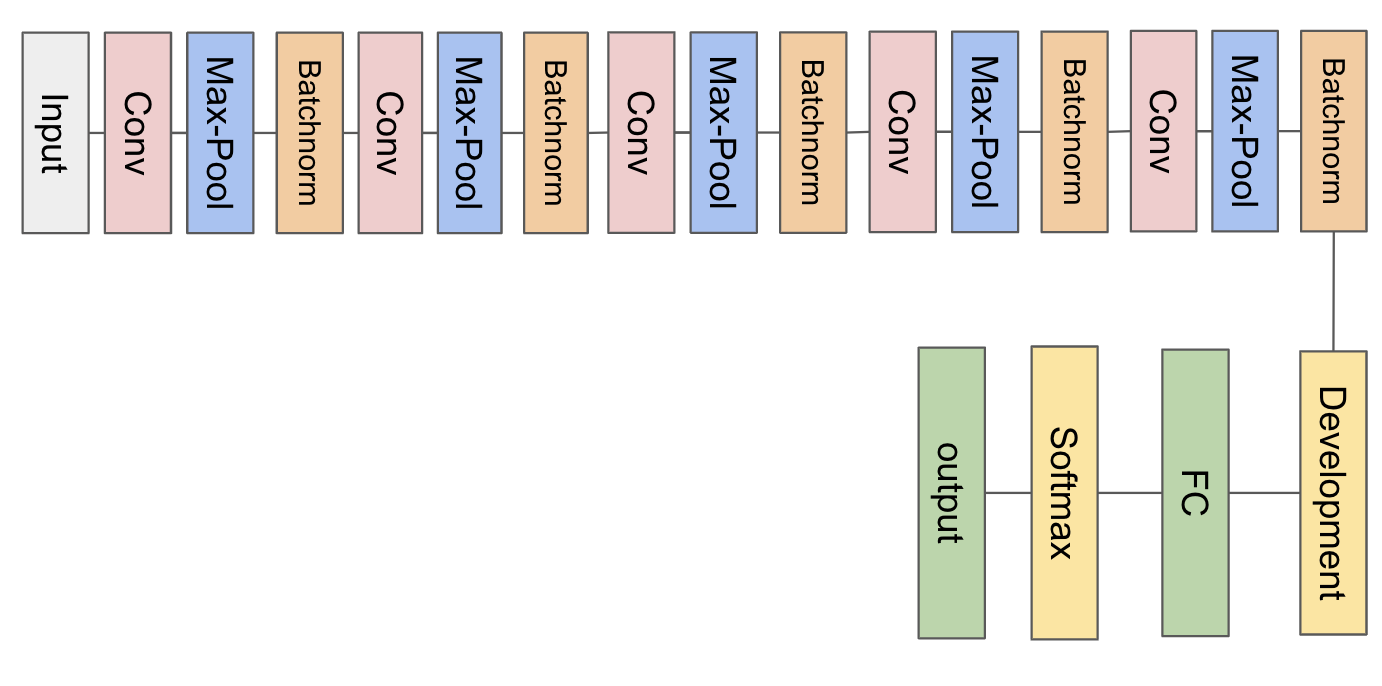} 
  \caption{DEV\_Model}
  \label{fig:example} 
\end{figure}

\section{Numerical Experiment Results}

\subsection{The Runtime Environment of the Program}

The research platform is hosted on an Alibaba Cloud server located in China, operating under the Linux 22.04 environment. We utilized Python 3.7 for programming and developed and trained deep learning models in the PyTorch 1.9.0 environment, complemented by the Path Signature library version 1.2.6.1.9.0. The system is powered by an Intel Xeon Platinum 8369B CPU, equipped with 8 cores and 16 threads, and a base frequency of 2.90GHz. The CPU features 384 KiB of L1d cache, 256 KiB of L1i cache, 10 MiB of L2 cache, and 48 MiB of L3 cache.

The Graphic Processing Unit (GPU) is an NVIDIA A100-SXM-80GB, operating with the driver version 460.91.03, compatible with CUDA 11.2, and houses 80GB of video memory, providing computational support for our deep learning algorithms.

\subsection{Hyperparameter Range} 

This study undertakes hyperparameter optimization, encompassing variables \texttt{CNN\_LayerNumber}, \texttt{LSTM\_LayerNumber}, \texttt{LSTM\_Number}, \texttt{L2\_Weight}, \texttt{CNN\_kernel}, \texttt{CNN\_stride}, \texttt{DEV\_Number}, \texttt{DNN\_Number}, and \texttt{CNN\_Channel}.

The hyperparameters, namely, \texttt{lr} (learning rate), \texttt{epoch} (number of epochs), and \texttt{batch\_size} (batch size), were selected based on machine learning model experience.

The range of these hyperparameters can be represented using this table.

\begin{tabular}{|c|c|}
\hline
Hyperparameter & Range \\
\hline
Learning Rate (\texttt{lr}) & \{0.001, 0.01\} \\
\hline
Number of Epochs (\texttt{epoch}) & \{100, 150, 300\} \\
\hline
Batch Size (\texttt{batch\_size}) & \{32, 64, 128\} \\
\hline
Number of CNN Layers (\texttt{CNN\_LayerNumber}) & [5, 10] \\
\hline
Number of CNN Channels (\texttt{CNN\_Channel}) & [6, 32] \\
\hline
CNN Kernel Size (\texttt{CNN\_kernel}) & [3, 12] \\
\hline
CNN Stride (\texttt{CNN\_stride}) & [1, 2] \\
\hline
Number of LSTM Layers (\texttt{LSTM\_LayerNumber}) & [1, 5] \\
\hline
LSTM Hidden Units (\texttt{LSTM\_Number}) & [12, 64] \\
\hline
L2 Regularization Weight (\texttt{L2\_Weight}) & [0, 0.05] \\
\hline
Path Development Layer Matrix Length (\texttt{DEV\_Number}) & [16, 32] \\
\hline
Number of DNN Hidden (\texttt{DNN\_Number}) & [16, 64] \\
\hline
\end{tabular}

Training every hyperparameter combination from this table would necessitate the training of nearly 6.1 billion models, an endeavor that is pragmatically unfeasible. Certain research initiatives employ methodologies such as random search and Bayesian optimization for the fine-tuning of hyperparameters\cite{bergstra2012random}\cite{snoek2012practical}. Nonetheless, this particular scholarly work opted for the application of the coordinate descent algorithm to achieve this objective, which follows a sequential approach, addressing and training each hyperparameter independently.

The memory requirements for the path development layer are generally higher compared to other networks. In experiments, it has been observed that even when using an A100 GPU, for a slightly longer time series, there is often a complete depletion of memory during training. Therefore, it is advisable to use small batches and a lower learning rate than usual.

Data augmentation was performed, so this paper did not use the focal loss as the loss function\cite{lin2020focal}. Instead, it employed the cross-entropy loss function and chose the Adam optimizer\cite{kingma2017adam}.

\subsection{Evaluation and Comparative Analysis of Model Performance} 

This research is devoted to developing a model for more accurately identifying which patients with atrial fibrillation (AF) actually do not need to take anticoagulants. While the current medical strategy can accurately identify all the AF patients who genuinely require drug therapy, it also results in a large majority being recommended for drug treatment.

To improve upon this, we have set two basic criteria for our model: firstly, if the model recommends that a patient does not need to take medication, it must ensure that the patient will not face a risk of stroke in the future due to not taking anticoagulants; secondly, the model should strive to identify as many AF patients as possible who genuinely do not need to take anticoagulants.

NPV is utilized to gauge the correctness of a diagnostic test in predicting a negative outcome. It is computed as:

\[
\text{NPV} = \frac{\text{TN}}{(\text{TN} + \text{FN})}
\]

Where:
\begin{itemize}
  \item \textbf{TN (True Negatives)}: The count of instances correctly predicted as negative.
  \item \textbf{FN (False Negatives)}: The count of positive instances incorrectly predicted as negative.
\end{itemize}

NPV provides the probability that a predicted negative outcome is indeed negative.

Specificity measures the ability of a diagnostic test to correctly identify negative outcomes. It can be formulated as:

\[
\text{Specificity} = \frac{\text{TN}}{(\text{TN} + \text{FP})}
\]

Where:
\begin{itemize}
  \item \textbf{TN (True Negatives)}: The count of instances correctly predicted as negative.
  \item \textbf{FP (False Positives)}: The count of negative instances incorrectly predicted as positive.
\end{itemize}

In the dataset, positive indicates the occurrence of a stroke in the future, while negative indicates the absence of a stroke. Based on meeting these fundamental criteria, we further propose two target indicators for the model: the NPV must be 1 to ensure no possibility of misdiagnosis; under the premise of achieving an NPV of 1, we will also work to increase the specificity of the model, aiming to more precisely identify those AF patients who do not require drug therapy.

Validation and test sets typically consist of 90 samples, with 17 being positive samples. For a general medical model, a Negative Predictive Value (NPV) of at least 99\% is required for use. Therefore, in our experiments, we selected an NPV of 1 to meet the predefined criteria.

\subsection{Training, Validation, and Testing of the Model} 

In our study, we partitioned the dataset into three distinct subsets: \(80\%\) serving as the training set, \(10\%\) designated as the test set, and the remaining \(10\%\) allocated for validation. The training regimen consisted of several epochs, each concluding with a validation phase to fine-tune the model parameters.

We devised a dynamic threshold selection mechanism to hone the model's Negative Predictive Value (NPV) and specificity. Since the model outputs two probability scores following softmax activation, we defined the positive class predictions through a meticulous exploration of various thresholds. Specifically, we scrutinized thresholds within a continuum from \(0\) to \(1\), ultimately selecting a threshold that ensured an NPV of \(1\) while concurrently maximizing specificity.

Upon the culmination of all epochs, we documented the model exhibiting the pinnacle of specificity on the validation set along with the corresponding threshold. This methodology not only underscores the theoretical efficacy of our model but also signals its viability for real-world applications. The chosen threshold and model are designed to maximize generalization ability.

The hyperparameters and training strategies for the comparison model are identical.

The error on the training set during a typical training process can be observed in Figure 4.11.

\begin{figure}[h] 
  \centering
  \includegraphics[width=1\textwidth]{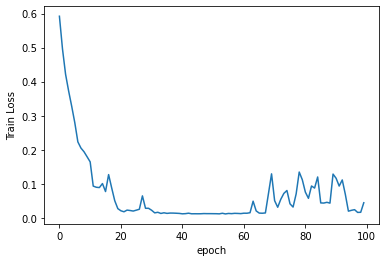} 
  \caption{Training Set Error}
  \label{fig:example} 
\end{figure}

The validation set error typically exhibits slightly larger fluctuations. A good hyperparameter selection is reflected in the fluctuations of the validation set error in Figure 4.12.

\begin{figure}[h] 
  \centering
  \includegraphics[width=1\textwidth]{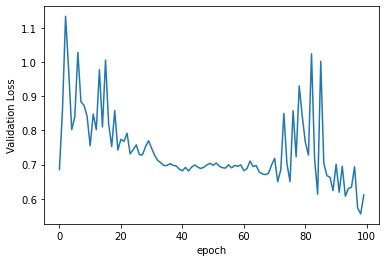} 
  \caption{Validation Set Error}
  \label{fig:example} 
\end{figure}

The specificity on the validation set for each epoch during training in a typical model, excluding rounds where specificity was 0, is depicted in Figure 4.13 and Figure 4.14.

\begin{figure}[h] 
  \centering
  \includegraphics[width=1\textwidth]{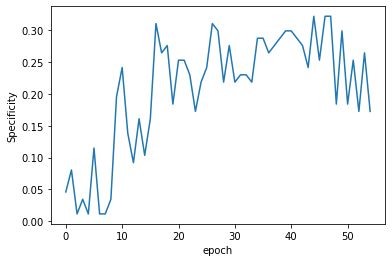} 
  \caption{Model's Specificity on the Validation Set}
  \label{fig:example} 
\end{figure}

\begin{figure}[h] 
  \centering
  \includegraphics[width=1\textwidth]{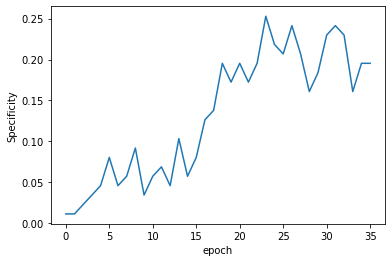} 
  \caption{Model's Specificity on the Validation Set}
  \label{fig:example} 
\end{figure}

After selecting hyperparameters, 2000 models were generated. They were evaluated on the validation set, and the best hyperparameters were chosen. Then, the selected model was tested on the test set, and the results are presented in \textbf{Table Model Hyperparameters and Performance Metrics Comparison}. The \texttt{lr}, \texttt{epoch}, \texttt{batch\_size}, and \texttt{DNN\_Number} in this context are chosen empirically.

\begin{table}[h]
\centering
\caption{Model Hyperparameters and Performance Metrics Comparison}
\hspace*{-1cm}
\begin{tabular}{|c|c|c|c|c|c|c|}
\hline
Hyperparameter& Model1 & Model2 & Model3 & Model4 & DEV\_Model & LSTM\_Model\\
\hline
\texttt{lr} & 0.001 & 0.001 & 0.001 & 0.001 & 0.001 & 0.001 \\
\hline
\texttt{epoch} & 100 & 150 & 150 & 150 & 150 & 150 \\
\hline
\texttt{batch\_size} & 32 & 32 & 32 & 32 & 32 & 128 \\
\hline
\texttt{CNN\_LayerNumber} & 5 & 6 & 8 & 8 & 10 & 10\\
\hline
\texttt{CNN\_Channel} & 6 & 12 & 20 & 32 & 12 & 12 \\
\hline
\texttt{CNN\_kernel} & 3 & 7 & 7 & 3 & 6 & 6 \\
\hline
\texttt{CNN\_stride} & 2 & 2 & 2 & 2 & 2 & 2 \\
\hline
\texttt{LSTM\_LayerNumber} & 1 & 2 & 1 & 2 & 0 & 2 \\
\hline
\texttt{LSTM\_Number} & 50 & 12 & 50 & 50 & 0 & 10 \\
\hline
\texttt{DEV\_Number} & 16 & 30 & 30 & 16 & 16 & 0 \\
\hline
\texttt{DNN\_Number} & 16 & 16 & 16 & 16 & 16 & 16 \\
\hline
\texttt{L2\_Weight} & 0.01 & 0 & 0.01 & 0.05 & 0.02 & 0 \\
\hline
\texttt{NPV} & 1 & 1 & 1 & 1 & 1 & 1 \\
\hline
\texttt{specificity} & 6.7\% & 9.1\% & 16.2\% & 30.6\% & 14.9\% & 2.7\% \\
\hline
\end{tabular}
\end{table}

In this study, we conducted an in-depth comparison and analysis of different feature extraction methods. We found that the model combining LSTM networks and Path Development Layers performed remarkably well, achieving a specificity of 30.7\%. In contrast, the model relying solely on Path Development Layers for feature extraction showed a slightly inferior performance, with a specificity of 14.9\%. Most notably, the model exclusively utilizing LSTM networks for feature extraction exhibited the poorest performance, with a specificity of only 2.3\%.

Furthermore, our research revealed the significant impact of the number of channels in convolutional layers on models employing Path Development Layers. An increase in the number of channels typically correlated with superior performance, as it facilitated the extraction of a more diverse set of features for subsequent learning. On the other hand, we observed that the matrix size within the Path Development Layer had a relatively limited impact on learning outcomes. This was attributed to the fact that the information captured by path signatures diminishes exponentially with each successive Picard iteration, all the information encapsulated within the path signatures can be located within the Path Development Layer\cite{lou2022path}. This may represent a strategy for explanation.

Furthermore, it can also be observed that deep convolutional layers tend to perform better compared to larger convolutional kernels. While both approaches can increase the receptive field, larger convolutional kernels come with a higher number of parameters, which may lead to overfitting. In contrast, deep convolutional layers have relatively fewer parameters. Additionally, deeper convolutional networks have the ability to learn more abstract features, aligning with our earlier findings regarding the effectiveness of deep neural networks.

\section{Future Research Directions}

\subsection{Strategies for Finding Optimal Hyperparameters} 

During our model development process, we discerned that the path development layer demands a significantly higher amount of GPU memory compared to other models. This characteristic has obligated us to reduce the batch size to accommodate the high memory demand, yet this reduction has engendered greater volatility during the training phase than is typical.

Another issue to highlight is the substantial number of models utilized in this research paper, making the task of identifying the optimal hyperparameters for each model notably time-consuming and complex. To streamline this process as much as possible, we resorted to making some hyperparameter choices based on prior experiences, deciding to maintain certain hyperparameters stable throughout the training period. However, we have to acknowledge the likelihood that these empirical choices might not necessarily offer the best solution to the problem at hand.

We also noticed that both the LSTM network and the path development layer exhibit extremely slow training speeds. This not only impacts the efficiency of model training but also signifies that verifying the correctness of a single hyperparameter will demand a prolonged duration. This undoubtedly augments the challenges we face in the stages of model development and optimization.

Therefore, in our future work, we plan to delve further into optimizing the methodology for hyperparameter selection, and to explore the possibility of accelerating the training process through methodological improvements, aiming to enhance efficiency while retaining high accuracy.

\subsection{Large Dataset} 

This paper conducts analysis and modeling based on a relatively small dataset of 871 atrial fibrillation patients provided by the Jiangsu Provincial People's Hospital. The limited sample size renders our models more prone to overfitting, a situation exacerbated by the fact that both LSTM networks and Path Development Layers generally involve a large number of parameters.

Expanding the sample size could be a viable solution to reduce overfitting and to enhance the generalization and stability of the models. In the medical field, ECG recordings frequently display long-term data correlations, warranting the exploration of more robust feature extraction strategies such as attention mechanisms\cite{NIPS2017_3f5ee243}\cite{xu2016show}. Although utilized in our research, attention mechanisms did not surpass the performance of convolutional layers. Generally, attention mechanisms require a more extensive sample size to perform optimally, suggesting that a larger dataset would allow for more effective feature extraction techniques, potentially yielding better results and correlations compared to convolutional layers. While these mechanisms did not outperform the convolutional layers in our research, it is envisaged that in future studies, attention mechanisms could potentially be integrated post the convolutional layers to perhaps foster better results and correlations\cite{brarda-etal-2017-sequential}\cite{song2017attend}.

One strategy to increase the dataset size compared to before is data augmentation. This paper has employed data augmentation, albeit to a somewhat limited extent. To further enhance data augmentation, the use of Generative Adversarial Networks (GANs) can be considered\cite{goodfellow2014generative}\cite{radford2016unsupervised}. GANs consist of two components – a generator and a discriminator. The generator aims to create convincing fake data to the point where the discriminator cannot differentiate between real data and the generated fake data. Through this, it can create new data points that are similar to those in the existing dataset yet novel\cite{10.1016/j.specom.2019.08.006}\cite{DBLP:conf/iclr/BrockDS19}. This network can also be incorporated into a model at the path development layer for predictions.

\subsection{Long-Term Effects and Interpretability} 

Upon conducting comprehensive dialogues with specialists from Jiangsu Province People's Hospital, we are able to acquire subsequent medical records of patients exhibiting atrial fibrillation within the dataset. This will facilitate our validation of the model's capability to accurately forecast the incidence probability and temporal occurrence of stroke in patients with atrial fibrillation. We intend to scrutinize the duration within which patients, deemed by the model as not necessitating medication, remain stroke-free. Should the outcomes prove affirmative, it would signify that the predictive capacity of the model extends beyond a one-year frame, potentially ensuring that patients remain medication-free for multiple years. A proficient performance from the model would corroborate its acquisition of pivotal characteristics.

The interpretability primarily centers on elucidating the rationale behind the selection of these hyperparameters. In convolutional neural networks, possessing a larger receptive field is generally associated with improved performance. Moreover, through in-depth research, it has been observed that models exhibiting superior results typically have time series consisting of several hundred time points post-convolution.

This intriguing phenomenon may bear a certain correlation with the gradient vanishing problem encountered by LSTM. Although LSTM are explicitly designed to alleviate the issues associated with gradient vanishing, they still demonstrate limitations in preserving the intricacy and nuance of data when tasked with processing particularly long sequences.

Such insights and observations derived from the operational characteristics of LSTM inevitably influence the strategies and decisions related to hyperparameter selection. The delicate balance between sequence length and memory retention in LSTM unveils critical factors that must be considered during the model development phase. By adopting and integrating this perspective into model development strategies, we anticipate refining the approaches to hyperparameter optimization more precisely.

\chapter{Conclusions}
\label{chapterlabel4}

This study underscores the pivotal role of electrocardiograms in predicting the likelihood of stroke and future medication requirements for atrial fibrillation patients in rural Jiangsu Province. We investigated patients who were recorded in the hospital but were reluctant to receive treatment, documenting their electrocardiograms and strokes in the following year. The results indicate that $83.6\%$ of the patients in the provided training set did not require medication. However, due to the high risk of stroke, the majority of patients are still recommended to undergo medication, even though this would reduce coagulation ability.

After evaluation by experts from Jiangsu Province People's Hospital, empirical evidence demonstrates that by solely substituting the Path Development Layer for Long Short-Term Memory networks (LSTM), a specificity of $14.9\%$ can be achieved. Furthermore, when combining the Path Development Layer with LSTM networks, the model’s specificity remarkably reaches $30.6\%$, significantly outperforming the model relying solely on LSTM networks, which has a specificity of $2.7\%$. This suggests that $2.31\%$ of atrial fibrillation patients could benefit from the LSTM network model, while a substantial $25.6\%$ could benefit from the model incorporating the Path Development Layer.

We also discovered that the Path Development Layer could replace LSTM networks in time series analysis and that its feature extraction capability is notably superior to existing methods. Moreover, the Path Development Layer can be integrated with LSTM networks within a deep learning framework to optimize feature extraction and learning processes in time series analysis.

In summary, this research unveils the considerable potential of using the Path Development Layer, either independently or in conjunction with LSTM networks, offering new avenues for future research. It demonstrates significant superiority in predicting the probability of stroke and individualized medication needs for atrial fibrillation patients, compared to the CHA2DS2-VASc strategy.

\phantomsection
\addcontentsline{toc}{chapter}{Appendices}

\appendix
\chapter{An Appendix About Stuff}
\label{appendixlabel1}
\begin{theorem}
Let \( K\subset \mathbb{R}^m\) be compact, \(\forall k \in \mathbb{N}\), polynomial of degree \( k \) are not dense in \( C(K) \) space
\begin{proof}
The set of all polynomials of degree \(n\) or less, denoted as \(P_n\), \(P_n \subset C(K)\), forms a vector space over the field of real numbers. This vector space can be defined as follows:

\[
P_n = \left\{ p(x) = a_0 + a_1 x + a_2 x^2 + \ldots + a_n x^n \mid a_i \in \mathbb{R}, \, i = 0,1,2,\ldots,n \right\}
\]

The dimension of this vector space is \(n+1\), and a basis for this space, known as the standard basis, is given by the set 

\[
\mathcal{B} = \{1, x, x^2, \ldots, x^n\}.
\]

In this basis, a polynomial \(p(x)\) in \(P_n\) can be uniquely represented as a linear combination of the basis vectors:

\[
p(x) = a_0 \cdot 1 + a_1 \cdot x + a_2 \cdot x^2 + \ldots + a_n \cdot x^n.
\]

Being a finite-dimensional subspace, \(P_n\) is also a complete subspace of \(C(K)\), meaning that it is closed and contains all its limit points. This property ensures that the convergence of a sequence of functions in \(P_n\) still results in a function that belongs to \(P_n\). \(e^x \notin P_n\), \(P_n\) is not a dense subspace in \(C(K)\).
\end{proof}
\end{theorem}

\section{Training, Validation, and Test}

In the contemporary context of machine learning methodologies, particularly within the realm of neural network architectures, the systematic division of data into training, validation, and test subsets is of pivotal importance. This tripartition is predicated on the necessity for precise adjustment of both model parameters and hyperparameters.

Here, it's worth explicitly clarifying what hyperparameters are. In the context of machine learning and neural networks, hyperparameters are parameters that are manually set before the learning process begins, as opposed to model parameters that are learned automatically during training\cite{Yang2020}. For example, in neural networks, weights and biases are learned through data and training algorithms\cite{Yu2020}. In contrast, hyperparameters remain constant throughout the training process. Common hyperparameters include the learning rate, batch size, number of hidden layers, and the number of nodes in each layer, among others.

The training subset is primarily employed for the iterative adjustment of intrinsic model parameters. Specifically, in the context of neural networks, weights and biases—denoted as 
\( \mathbf{W_h^*} \),\( \mathbf{b_h^*} \),\( \mathbf{W_o^*} \),\( \mathbf{b_o^*} \)—are optimized using techniques such as stochastic gradient descent or its variants. The objective of this process is to minimize a predefined loss function, thereby refining the model's predictive or inferential capabilities.

In contrast, the validation subset plays a significant role in the heuristic calibration of hyperparameters. One notable aspect is the determination of the optimal matrix dimensions within the neural network architecture. By evaluating the model's performance across a multidimensional hyperparameter space, one can empirically discern which matrix size produces the highest performance metrics, such as accuracy, F1 score, or area under the receiver operating characteristic curve (AUC-ROC)\cite{Bergstra2011}\cite{Davis2006}.

Once the optimal hyperparameters have been determined, a common practice is to retrain the model on the entire training dataset, which comprises both the original training and validation subsets. This step aims to make full use of the maximum available data, thereby potentially enhancing the model's statistical robustness.

Lastly, assessing the model on a separate test set serves as an epistemic litmus test for its generalizability. This evaluation provides an unbiased appraisal of the model's performance on novel, unseen data, thus validating its capability for reliable predictions or inferences in real-world applications.

It is important to note that the effectiveness of this data partitioning paradigm rests on the assumption of Independent and Identically Distributed (IID) data across all subsets. This presupposition posits that each data sample is independently drawn from a consistent underlying distribution, thereby ensuring the statistical validity of the training, validation, and testing processes. If this IID assumption is violated, alternative methodologies such as domain adaptation or transfer learning may be required to ensure the model's generalizability and robustness.

\section{No Free Lunch Theorem, Overfitting}

The approximation theorems relevant to neural networks establish the foundational viability of these algorithms for approximating a broad class of functions. However, such theoretical frameworks often fall short in providing specific guidance on the optimal architecture or complexity of the model. Notably, the true underlying function to be approximated is usually unknown in practical applications, underscoring the importance of domain expertise in both model selection and interpretation\cite{Janiesch2021}.

In a typical machine learning experiment, for each instance of sampling, an observation is derived from the stochastic process \( Y = X^2 + \epsilon \), where \( \epsilon \) is an i.i.d. Gaussian random variable. Utilizing a parametric model of the form \( f(x) = ax^2 \) to fit a dataset of \( N = 10,000 \) independently and identically distributed observations, the model is anticipated to approximate the ground truth function \( Y = X^2 \). The expected bias contributing to the \( L_2 \) loss function predominantly arises from the Gaussian-distributed noise term \( \epsilon \).

Models of such simplistic structure generally exhibit higher bias but lower variance. They are adept at capturing the global trend of the data while neglecting the idiosyncratic noise inherent in each individual observation. In contrast, using a highly complex model, such as a polynomial of degree 10,000, enables the model to perfectly interpolate each data point in the training set, including the Gaussian noise \( \epsilon \).

This scenario exemplifies a prototypical case of overfitting. The model manifests an extremely low bias in the training set yet suffers from an exorbitant amount of variance owing to its complexity\cite{NIPS2000_059fdcd9}. Consequently, its generalization capability on novel, out-of-sample data, such as a validation dataset, is anticipated to be poor, leading to a substantially elevated prediction error\cite{shalevshwartz2014regularization}.

Overfitting refers to the phenomenon where a model exhibits low bias but high variance on the training dataset\cite{shalevshwartz2014regularization}.

Consider a simplistic scenario where the target function is \(y = x^2\). Even in this ostensibly straightforward case, the inherent sparsity of a real-world training set---typically unable to comprehensively sample the domain of \(\mathbb{R}\)---can lead to model ambiguity. For example, if the training data comprises only the points \((0,0)\) and \((1,1)\), models based on functions such as \(y = x\), \(y = x^2\), and \(y = x^4\) would all yield perfect fits to the training data. Yet their generalization performance on an independent validation set could vary dramatically, illustrating the risk of overfitting when the model complexity is unnecessarily high.

This begs the question: if the true underlying data-generating process were to indeed correspond to a higher-order polynomial, would such complexity still constitute overfitting? The answer is nuanced. In instances where independent validation or out-of-sample testing confirms the model's efficacy, a more complex model could, in fact, be justifiable. This highlights the importance of model validation techniques and domain-specific insights in navigating the trade-offs between underfitting and overfitting, and illustrates the limitations of relying solely on theoretical guarantees such as approximation theorems.

In summary, while approximation theorems provide a theoretical backbone for neural networks, they do not obviate the need for empirical validation and nuanced judgment, particularly when the true function governing the data is unknown or complex.

The "No Free Lunch" (NFL) theorem complements this perspective by emphasizing that no single algorithm—whether a neural network or some other model—is universally optimal for all possible data-generating processes or loss functions\cite{10.1162/neco.1996.8.7.1341}. In other words, while approximation theorems may give us confidence in the expressive power of neural networks to approximate a wide range of functions, they do not offer any guarantees about the optimality of neural networks for any specific problem\cite{Wolpert1997}.

\begin{theorem}[No Free Lunch Theorem]\cite{Wolpert1997}
Let \( \mathcal{X} \) be the input space, \( \mathcal{Y} \) be the output space, \( \mathcal{X} \) and \( \mathcal{Y} \) are finite sets, and \( \mathcal{F} \) be the set of target functions defined on \( \mathcal{X} \) and taking values in \( \mathcal{Y} \). Let \( \mathcal{D} \) be the set of datasets, where each dataset \( D \) consists of sample pairs from \( \mathcal{X} \) and \( \mathcal{Y} \).

Let \( \mathcal{A} \) be a set of machine learning algorithms, where each algorithm \( A \) learns a mapping \( f_A: \mathcal{X} \to \mathcal{Y} \) by training on a dataset \( D \in \mathcal{D} \), \(A:\mathcal{D} \to \mathcal{F}, A(D)=f_A\)

Furthermore, let \( L: \mathcal{Y} \times \mathcal{Y} \to \mathbb{R} \) be a loss function.

Let \( \mu \) be a probability measure defined on \( \mathcal{F} \). In particular, we consider \( \mu \) as a uniform measure across \( \mathcal{F} \), i.e., each target function \( f \) is equally likely.

Then, \(\forall  A_1, A_2 \in \mathcal{A}, D \in \mathcal{D}, x\in \mathcal{X} \), we have:
\[
\mathbb{E}_{f \sim \mu}[L(f(x), f_{A_1}(x))] = \mathbb{E}_{f \sim \mu}[L(f(x), f_{A_2}(x))]
\]

Here, \( \mathbb{E}_{f \sim \mu}[L(f(x), f_A(x))] \) represents the expected loss of the mapping \( f_A \) learned by algorithm \( A \) across all possible target functions \( f \), under the measure \( \mu \).
\end{theorem}

Overfitting is the phenomenon where a model's performance does not necessarily improve with an increase in parameters, and it may even deteriorate due to the model fitting noise in the data too closely. The "No Free Lunch Theorem" states that, without a specific understanding of a problem domain, the performance of a model is as likely to be poor as it is to be good. In other words, it suggests that there is no universally superior model that performs well across all types of problems\cite{shalevshwartz2014biascomplexity}.

In machine learning, strategies commonly used to address the issue of overfitting typically include the following:

\begin{enumerate}[label=\arabic*.]
    \item \textbf{Increase Training Data}: Augmenting the volume of training data can assist the model in acquiring a more accurate representation of the underlying data distribution, thereby reducing the likelihood of overfitting.

    \item \textbf{Model Simplification}: Employ simpler model architectures, such as reducing the depth of neural networks or limiting the depth of decision trees, to mitigate model complexity.

    \item \textbf{Regularization}: Control the magnitude of model parameters by introducing additional penalty terms, such as L1 and L2 regularization\cite{51791361-8fe2-38d5-959f-ae8d048b490d}\cite{hoerl2000ridge}.

    \item \textbf{Cross-Validation}: Utilize cross-validation techniques to evaluate model performance rigorously, facilitating the detection and prevention of overfitting\cite{10.1214/09-SS054}.

    \item \textbf{Feature Selection}: Opt for feature selection methods to retain only the most pertinent features, thereby diminishing model complexity\cite{salahat2017recent}.

    \item \textbf{Early Stopping}: Monitor model performance during training, halting the process as soon as performance on the validation dataset begins to deteriorate, thus averting overfitting.

    \item \textbf{Ensemble Methods}: Deploy ensemble learning techniques, such as Random Forests or Gradient Boosting Trees, to mitigate the risk of overfitting.

    \item \textbf{Data Augmentation}: Enhance data diversity by applying random transformations to the training dataset, thereby reducing overfitting tendencies\cite{Shorten2019}.

    \item \textbf{Hyperparameter Tuning}: Adjust hyperparameters like learning rates and regularization strengths to optimize generalization performance\cite{maclaurin2015gradientbased}.
\end{enumerate}

\section{Feature Extraction}

The feature extraction process is a mapping from the input space \( \mathcal{X} \) to the machine learning model input space \( \mathcal{X}_1 \).At this point, the functions \(f\) in the machine learning model's hypothesis space \( \mathcal{H} \) are all mappings from the machine learning model input space \( \mathcal{X}_1 \) to output space \( \mathcal{Y} \).Generally speaking, \( \mathcal{X} \) is a subset of the space \( \mathbb{R}^n \) and \( \mathcal{X}_1 \) is a subset of the space \( \mathbb{R}^{n_1} \), where \( n_1 < n \).

Feature extraction refers to the process of extracting or transforming useful features from raw data to better describe, understand, and process data in machine learning and data analysis tasks\cite{guyon2003introduction}. These features can represent key information in the dataset or characteristics that are helpful in solving specific problems.

Feature extraction does not increase the amount of information.The central goal of feature extraction is to identify and extract the most critical and information-rich components from the initial dataset, not to increase the overall volume of information. Through this analytical process, it is possible to reduce the dimensionality and complexity of the data, thereby facilitating more efficient model training and potentially enhancing the model's predictive accuracy\cite{hira2015review}.

Feature extraction assists in the following aspects:

\begin{itemize}
    \item \textbf{Noise Reduction and Exclusion of Irrelevant Information}: This step helps to focus on the most critical features of the dataset by filtering out noise and other irrelevant information, thereby enhancing data purity and relevance.
   
    \item \textbf{Dimensionality Reduction}: By reducing the dimensionality of the dataset, it helps decrease the complexity of the model and reduces the risk of overfitting, a common problem in the domain of machine learning where the model learns excessively from the noise present in the training data.
   
    \item \textbf{Accelerating Model Training and Prediction}: Efficient feature extraction techniques can significantly speed up the training and prediction phases of model development, enhancing the operational efficiency of the model.

    \item \textbf{Enhanced Data Visualization}: By focusing on the most pertinent features, it enables more effective visualization of data and model outcomes, which is vital for interpreting the data and drawing accurate conclusions based on the model's output.
\end{itemize}

\section{Fourier and Wavelet Transform in Signal Processing}

Fourier Transform and Wavelet Transform are two pivotal analysis tools in signal processing, each providing unique insights into the characteristics of signals in both the frequency and time domains.

\subsection*{Fourier Transform}

The fundamental principle of the Fourier Transform is to convert a signal $f(t)$ from the time domain into the frequency domain, revealing the periodic characteristics of the signal. The expression for the Fourier Transform is:
\begin{equation}
F(k) = \int_{-\infty}^{\infty} f(t)e^{-2\pi ikt}dt
\end{equation}

The Fourier Transform excels at analyzing periodic and stationary signals. However, its performance diminishes when applied to non-stationary signals and those containing abrupt changes. This limitation arises because the basis functions of the Fourier Transform are sine and cosine functions, which are periodic across the entire real number domain, making them ill-suited for capturing local characteristics of signals, such as abrupt changes.

\subsection*{Wavelet Transform}

The Wavelet Transform introduces wavelet functions, which can be localized in the time domain, as basis functions. By scaling and translating, the Wavelet Transform facilitates the local analysis of signals in both the time and frequency domains. The expression for the Wavelet Transform is:
\begin{equation}
W_f(a, b) = \frac{1}{\sqrt{|a|}} \int_{-\infty}^{\infty} f(t) \psi^*\left(\frac{t - b}{a}\right)dt
\end{equation}

Here, $a$ and $b$ are the scaling and translation factors of the wavelet function, controlling its scale and position. This adaptability enables the Wavelet Transform to capture the characteristics of signals across different time scales and frequency bands, making it particularly suitable for analyzing non-stationary signals and abrupt changes.

In the context of electrocardiogram (ECG) signal processing, where the signal contains a plethora of non-stationary and abrupt information, the Wavelet Transform, due to its ability to localize in both time and frequency, has proven to be an effective method for noise reduction and feature extraction. Utilizing the Wavelet Transform allows for the elimination of noise while preserving the essential characteristics of the signal, thereby providing a clearer and more accurate foundation for subsequent ECG analysis and diagnosis.

\section{Attention Mechanism}

The attention mechanism is a technique used in machine learning to enable the model to focus selectively on different parts of the input sequence, assigning different importance or weights to different input elements based on their relevance to the task at hand.

\begin{enumerate}
    \item \textbf{Defining Query, Key, and Value (Parameter Learning):}
    
    For each input position $i$, we have:
    \begin{itemize}
        \item \textbf{Query $Q_i$:} A learned representation indicating the extent to which the model should attend to other positions. Calculated as $Q_i = W_qX_i$.
        \item \textbf{Key $K_i$:} A learned representation indicating how each position should be attended to. Calculated as $K_i = W_kX_i$.
        \item \textbf{Value $V_i$:} Represents the actual information content at each position. Calculated as $V_i = W_vX_i$.
    \end{itemize}
    
    \item \textbf{Computing Attention Scores (Intermediate Calculation):}
    
    The attention score between position $i$ and every other position $j$ is computed as the inner product of the Query and Key vectors, followed by scaling:
    \[ \text{score}(Q_i, K_j) = \frac{Q_i \cdot K_j^T}{\sqrt{d_k}} \]
    where $d_k$ is the dimension of the Key vector.
    
    \item \textbf{Normalization and Weight Calculation (Softmax Application):}
    
    For each position $i$, the attention weights for all positions $j$ are computed using the softmax function:
    \[ w_{ij} = \frac{\exp(\text{score}(Q_i, K_j))}{\sum_{j=1}^{n} \exp(\text{score}(Q_i, K_j))} \]
    
    \item \textbf{Weighted Aggregation (Context Vector Calculation):}
    
    The output at each position $i$ is the weighted sum of all Value vectors, representing the context information:
    \[ O_i = \sum_{j=1}^{n} w_{ij}V_j \]
    \item \textbf{Multi-Head Attention and Final Output (Output Aggregation):}
    
    Multi-head attention uses multiple sets of Query, Key, and Value weight matrices to produce multiple sets of output vectors\cite{LeNguyen2019}. These vectors are then concatenated and linearly transformed to produce the final output:
    \[ O_{\text{final}} = W_o\left(\bigoplus_{h=1}^{H} O_h\right) \]
    where $H$ is the number of heads, $O_h$ is the output of each head, and $W_o$ is a learned weight matrix.
\end{enumerate}

\chapter{Colophon}
\label{appendixlabel3}
\textit{The machine learning network model in this paper utilized the ML Visuals}


 

\bibliography{example}

\end{document}